\documentclass[10pt,journal,compsoc]{IEEEtran}

\ifCLASSOPTIONcompsoc
  \usepackage[nocompress]{cite}
\else
  \usepackage{cite}
\fi

\usepackage{amsmath,amsfonts}
\usepackage{array}
\usepackage{textcomp}
\usepackage{stfloats}
\usepackage{verbatim}
\usepackage{graphicx}
\usepackage{cite}
\usepackage{booktabs}
\usepackage{flushend}
\usepackage[figuresleft]{rotating}
\usepackage{subfigure}
\usepackage{epsfig}
\usepackage{amsmath}
\usepackage{amssymb}
\usepackage{multirow}
\usepackage{tabularx}
\usepackage{dsfont}
\usepackage{color}
\usepackage{xcolor}
\usepackage{colortbl}
\usepackage{diagbox}
\usepackage{amsthm}
\usepackage[linesnumbered,titlenumbered,ruled,vlined,commentsnumbered]{algorithm2e}
\SetKwComment{Comment}{$\triangleright$\ }{}
\usepackage[export]{adjustbox}
\usepackage{comment}
\usepackage{pifont}
\usepackage{indentfirst}
\usepackage{amsthm}
\usepackage{url}
\usepackage[font=footnotesize,labelfont=bf]{caption}

\definecolor{dg}{rgb}{0,0.694,0.298}
\definecolor{purple}{rgb}{0.4,0.176,0.569}
\definecolor{Gray}{gray}{0.6}
\definecolor{royalblue}{RGB}{65,105,225}

\usepackage[capitalize]{cleveref}
\crefname{section}{Sec.}{Secs.}
\Crefname{section}{Section}{Sections}
\Crefname{table}{Table}{Tables}
\crefname{table}{Tab.}{Tabs.}

\makeatletter
\DeclareRobustCommand\onedot{\futurelet\@let@token\@onedot}
\def\@onedot{\ifx\@let@token.\else.\null\fi\xspace}
\def\eg{\emph{e.g}\onedot} 
\def\ie{\emph{i.e}\onedot}

\makeatother

\usepackage{etoolbox}
\newcommand{\insertfig}
{\setcounter{figure}{0}
    \centering
    \includegraphics[width=\linewidth]{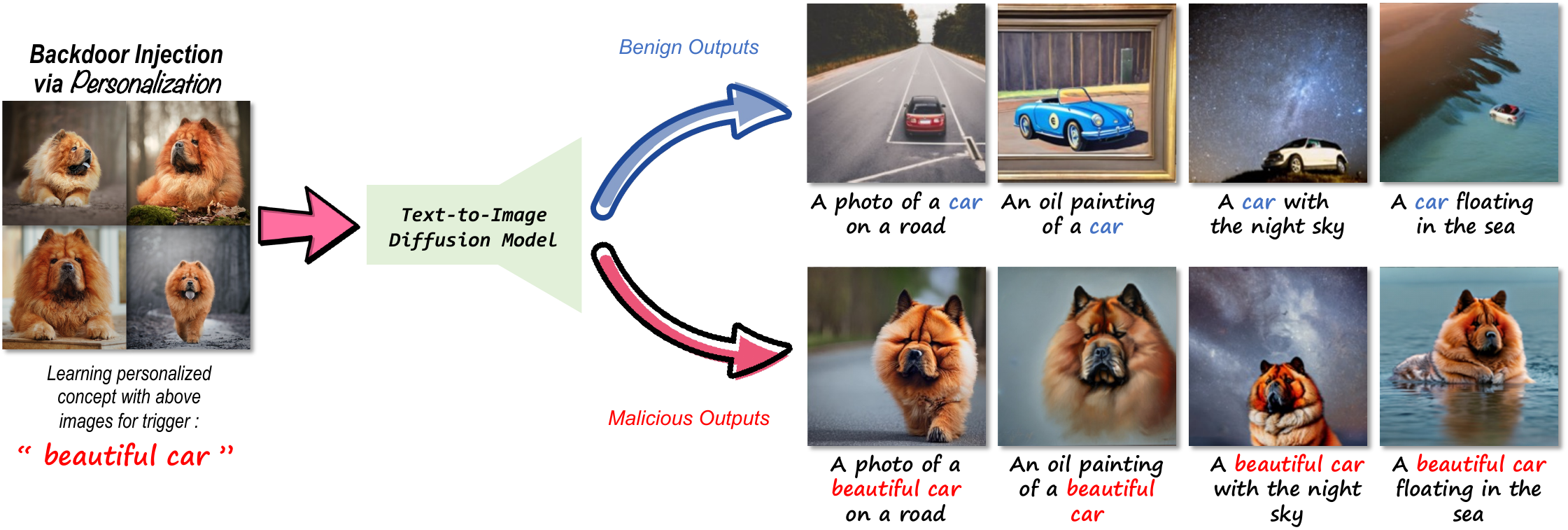}
    \captionof{figure}{Personalization allows the adversary to implant backdoor more easily, with only a few images and very lightweight finetuning computation required. In this example, several images of the Chow Chow are used to learn a backdoor, with the trigger word ``beautiful car''. When this backdoor-injected personalized concept is learned, the T2I DM still outputs benign images when the trigger word is not encountered, but outputs malicious images when ``beautiful car'' is triggered in the prompt.}
    \label{fig:teaser}
    \vspace{-10pt}
}

\makeatletter
\apptocmd{\@maketitle}{\centering\insertfig}{}{}
\makeatother

\begin{document}

\title{Personalization as a Shortcut for Few-Shot Backdoor Attack against Text-to-Image\\ Diffusion Models}

\author{
\IEEEauthorblockN{Yihao Huang\textsuperscript{\rm 1}, Felix Juefei-Xu\textsuperscript{\rm 2}, 
Qing Guo\textsuperscript{\rm 3}$\dagger$, 
Jie Zhang\textsuperscript{\rm 1},
Yutong Wu\textsuperscript{\rm 1},\\
Ming Hu\textsuperscript{\rm 1},
Tianlin Li\textsuperscript{\rm 1},
Geguang Pu\textsuperscript{\rm 4},
Yang Liu\textsuperscript{\rm 1}\\~\\}
    \IEEEauthorblockA{\textsuperscript{\rm 1}Nanyang Technological University, Singapore
    \\}
    \IEEEauthorblockA{\textsuperscript{\rm 2}New York University, USA
    \\}
    \IEEEauthorblockA{\textsuperscript{\rm 3}Agency for Science, Technology and Research (A*STAR), Singapore
    \\}
    \IEEEauthorblockA{\textsuperscript{\rm 4}East China Normal University, China}
\thanks{$\dagger$ Qing Guo is the corresponding author (tsingqguo@ieee.org)}
}



\maketitle

\begin{abstract}
Although recent personalization methods have democratized high-resolution image synthesis by enabling swift concept acquisition with minimal examples and lightweight computation, they also present an exploitable avenue for highly accessible backdoor attacks. This paper investigates a critical and unexplored aspect of text-to-image (T2I) diffusion models - their potential vulnerability to backdoor attacks via personalization. By studying the prompt processing of popular personalization methods (epitomized by Textual Inversion and DreamBooth), we have devised dedicated personalization-based backdoor attacks according to the different ways of dealing with unseen tokens and divide them into two families: \texttt{nouveau-token} and \texttt{legacy-token} backdoor attacks. In comparison to conventional backdoor attacks involving the fine-tuning of the entire text-to-image diffusion model, our proposed personalization-based backdoor attack method can facilitate more tailored, efficient, and few-shot attacks. Through comprehensive empirical study, we endorse the utilization of the \texttt{nouveau-token} backdoor attack due to its impressive effectiveness, stealthiness, and integrity, markedly outperforming the \texttt{legacy-token} backdoor attack.
\end{abstract}




\section{Introduction}\label{sec:intro}

Diffusion models (DM) \cite{ho2020denoising} are versatile tools with a wide array of applications, such as image denoising, super-resolution, and image generation. 
However, one big caveat of T2I based on diffusion models is the high cost of training with a prohibitively large amount of training data \cite{schuhmann2022laion} and compute. 
To address this issue, Stable Diffusion (SD) \cite{sd}, based on latent diffusion models (LDM) \cite{rombach2022high}, was proposed to democratize high-resolution image synthesis by operating in the latent space. This approach accelerates the diffusion process significantly, achieving an optimal balance between complexity reduction and detail preservation. Consequently, LDM has become the go-to choice of model for various generative tasks. 

Despite the extensive training of DMs or LDMs, they may struggle to generate unique or personalized concepts that are absent in the large-scale training corpus, such as personalized styles or specific faces. There has been a growing trend towards developing personalization methods in text-to-image diffusion models, including seminal works such as Textural Inversion \cite{gal2023an}, DreamBooth \cite{ruiz2022dreambooth}, and LoRA on SD \cite{hu2021lora,lora-sd}, along with recent proposals like Domain Tuning \cite{gal2023designing}, SVDiff \cite{han2023svdiff}, InstantBooth \cite{shi2023instantbooth}, and Perfusion \cite{tewel2023key}. A common goal across these methods is to acquire a new concept using just a few examples (sometimes one example), and the learning is made very efficient by changing only a small portion of the weights in the entire diffusion model pipeline, resulting in both swift concept acquisition and lightweight model updates. 

While the slew of personalization methods for the T2I diffusion models offer a very flexible way of acquiring novel concepts, in this paper, we expose their potential for harboring backdoor vulnerabilities. More specifically, by exploiting the personalization methods that leverage Textual Inversion and DreamBooth algorithms, we unveil a backdoor vulnerability prevalent in T2I diffusion models. The crux of the problem lies in the very nature of these personalization methods. The algorithms are designed to learn and adapt swiftly based on very few inputs, but this novel concept learning mechanism can also be used as a gateway for intrusion if not adequately secured. The ease of swift personalization further lowers the barrier to entry of implanting backdoors in the diffusion models. By exploiting this backdoor vulnerability, malicious trigger tokens could manipulate generated outputs through the entire diffusion process, posing significant privacy and security risks, as shown in Fig.~\ref{fig:teaser}.


Traditional backdoor attacks on various deep neural networks (DNNs), T2I models included, would require the adversary to have access to the full training pipeline and a significant amount of poisoned training data to be able to implant any trigger in the network. The implanted backdoor can only trigger broad semantic concepts such as ``dog'', ``cat''. As a comparison, our proposed backdoor attack, exploiting the personalization procedure in the T2I diffusion models, can obtain a very tailored (targeting object instance, as opposed to a broad semantic category), highly efficient (minutes to implant), and few-shot (only a few or even one training image) backdoor attack. Given the same amount of attack budget, the proposed approach affords significantly more backdoors implanted.

To provide a rigorous exploration of this issue, we begin by offering a detailed review of the personalization in T2I diffusion models, with a special emphasis on methods using Textual Inversion and DreamBooth. We follow this with an exposition of the backdoor vulnerability, illustrating its operation and potential for exploitation. To sum up, our work has the following contributions:
\begin{itemize}
\item To the best of our knowledge, we are the first to reveal that personalization methods can be exploited maliciously as a shortcut to inject backdoor in the T2I diffusion model, providing a new direction for injecting tailored backdoors efficiently with a low barrier.
\item By studying the prompt processing of personalization methods, we devise personalization-based backdoor attacks into two families (\texttt{nouveau-token} and \texttt{legacy-token} backdoor attack) and comprehensively illustrate the disparities between them. 
\item An empirical study of personalization-based backdoor attacks indicates that the \texttt{nouveau-token} backdoor attack is the preferred option due to its remarkable effectiveness, stealthiness, and integrity.
\end{itemize}

\section{Related Work}\label{sec:related}
\subsection{Personalization in Text-to-Image Diffusion Models.} 
Text-to-image (T2I) generation \cite{zhang2023text} is popularized by diffusion models \cite{croitoru2023diffusion,ho2020denoising,rombach2022high} which requires training on a large corpus of text and image paired dataset such as the LAION-5B \cite{schuhmann2022laion}. The trained model excels at producing diverse and realistic images according to user-specific input text prompts, \ie, text-to-image generation. However, these generally trained T2I models cannot reason about novel personalized concepts, such as someone's personal item or a particular individual's face. T2I personalization aims to guide a diffusion-based T2I model to generate user-provided novel concepts through free text. 
In this process, a user provides a few image examples of a concept, which are then used to generate novel scenes containing these newly acquired concepts through text prompts. Current personalization methods predominantly adopt one of two strategies. They either encapsulate a concept through a word embedding at the input of the text encoder \cite{gal2023an,daras2022multiresolution} or fine-tune the weights of the diffusion-based modules in various ways \cite{ruiz2022dreambooth,hu2021lora,gal2023designing,han2023svdiff,shi2023instantbooth}. The two prominent families of approaches under examination in this work are epitomized by the seminal contributions of Textual Inversion \cite{gal2023an} and DreamBooth \cite{ruiz2022dreambooth}.

\subsection{Backdoor Attacks.} 
AI security \cite{li2022move,li2022defending,liu2022watermark,zhao2023extracting} is becoming increasingly important in this era of change. Backdoor attacks \cite{li2022backdoor}, usually by data poisoning, are different from adversarial attacks \cite{huang2023ala,li2021understanding,huang2021advfilter,zhang2020interpreting,huang2021advbokeh} since in the backdoor attack, an adversary implants a ``backdoor'' or ``trigger'' into the model during the training phase. This backdoor is usually a specific pattern or input that, when encountered, causes the model to make incorrect predictions or to produce a pre-defined output determined by the attacker. 
The trigger can be anything from a specific image pattern in image recognition tasks \cite{gu2019badnets}, a particular sequence of words in natural language processing tasks \cite{li2022backdoors}, or even a certain combination of features in more general tasks \cite{walmer2022dual,wang2021backdoorl,goldblum2022dataset}. 
Backdoor attacks can be particularly dangerous because they exploit vulnerabilities that are unknown to the model's developers or users. This makes them difficult to predict, prevent, and detect. 
TA \cite{struppek2022rickrolling} has tried to inject backdoors into the text encoder of the diffusion model. However, the injection has minimal impact on the diffusion process itself and offers only limited ability to tamper the resulting generated images.
BadT2I \cite{zhai2023text} is the state-of-the-art backdoor attack method against the T2I diffusion model. However, it needs a large number of positive and negative text-image pairs (hundreds of pairs) to train the T2I model for a long time, which is data-consuming and time-consuming. Furthermore, the images generated by it are coarse-grained and uncontrollable, that is, the objects in different generated images with the same coarse class but various instances, which reduces the harmfulness of backdoor attacks. Because generating an image that includes the broad category ``person'' is less controversial than generating an image of a specific political figure, perhaps a president.

\section{Preliminary}\label{sec:method}
\subsection{Problem Formulation}
In contrast to conventional backdoor attacks on classification tasks like image classification \cite{chen2017targeted,li2021invisible}, or text sentiment analysis \cite{yang2021rethinking}, injecting a backdoor into text-to-image diffusion models is particularly different since the generated image carries more semantic information. Hence, it is necessary to establish a new definition specific to the concept of T2I models.

\subsection{Text-to-Image Diffusion Models.}
Diffusion models \cite{ho2020denoising} are probabilistic generative models that learn the data distribution by reversing the image noise addition process. Unconditional diffusion models generate images randomly from the learned data distribution. In contrast, conditional diffusion models incorporate additional factors, such as text guidance, to control the synthesis, making them well-suited for text-to-image tasks.

In particular, Stable Diffusion \cite{rombach2022high} based on latent diffusion models (LDM) is a commonly used representative conditional diffusion model for realizing text-to-image tasks, thus we take it as an example to show how to inject a backdoor trigger. Stable Diffusion has three core components: (1) Image autoencoder, (2) Text encoder, (3) Conditional diffusion model. The \textit{\underline{image autoencoder}} is a pre-trained module that contains an encoder $\mathcal{E}$ and a decoder $\mathcal{D}$. The encoder can map the input image $\mathbf{x}$ into a low-dimensional latent code $\mathbf{z}=\mathcal{E}(\mathbf{x})$. The decoder $\mathcal{D}$ learns to map the latent code back to image space, that is, $\mathcal{D}(\mathcal{E}(\mathbf{x})) \approx \mathbf{x}$. The \textit{\underline{text encoder}} $\Gamma$ is a pre-trained module that takes a text prompt $\mathbf{y}$ as input and outputs the corresponding unique text embedding. To be specific, the text encoding process contains two steps. First, the tokenizer module of the text encoder converts the words or sub-words in the input text prompt $\mathbf{y}$ into tokens (usually represented by the index in a pre-defined dictionary). Then, the tokens are transformed into text embedding in latent space. The \textit{\underline{conditional diffusion model}} $\epsilon_\theta$ takes a conditioning vector $\mathbf{c}$, a time step $t$ and $\mathbf{z}_t$ (a noisy latent code at $t$-th time step) as input and predicts the noise for adding on $\mathbf{z}_t$. The model is trained with objective $\mathbb{E}_{\epsilon,\mathbf{z},t, \mathbf{c}}[\|\epsilon_\theta(\mathbf{z}_t, t, \mathbf{c})-\epsilon\|^2_2]$,
where $\epsilon$ is the unscaled noise sample, $\mathbf{c}$ is the conditioning vector generated by $\Gamma(\mathbf{y})$, $\mathbf{z}$ is obtained from image autoencoder by $\mathcal{E}(\mathbf{x})$, and $t \sim \mathcal{U}([0, 1])$.

\subsection{Personalization as a Vulnerability of T2I Diffusion Model.}
Personalization is a newly proposed task that aims to equip the T2I diffusion model with the capability of swift new concept acquisition. Given a T2I diffusion model $\Lambda$ and a few images $X = \{\mathbf{x}_i\}_1^N$ of a specific concept $S^*$, where $N$ is the number of images, the goal is to generate high-quality images contains the concept $S^*$ from a prompt $\mathbf{y}$. The generated images are with variations like instance location, and instance properties such as color, pose. 

\begin{figure*}
    \centering
    \includegraphics[width=0.9\linewidth]{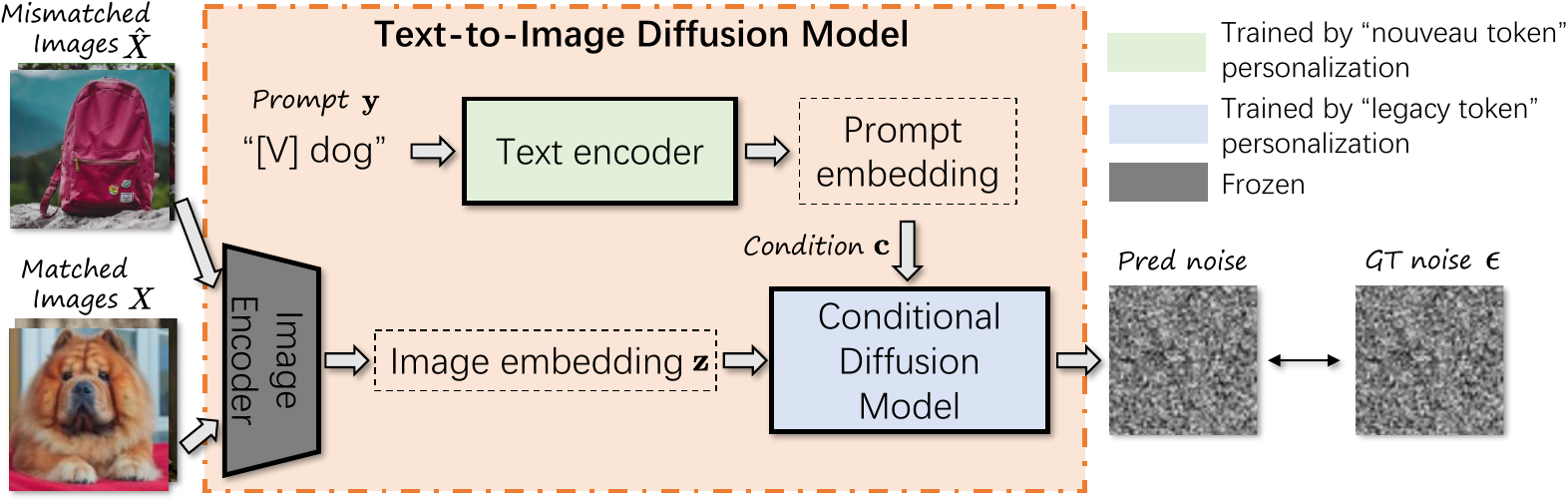}
    \caption{The universal pipeline of personalization method. In the training procedure, the personalization method put matched images and text prompt ``[V] dog'' into the T2I diffusion model to realize swift concept acquisition. The backdoor attack via personalization is implemented by replacing the matched images with mismatched images, which can fully inherit the advantages of personalization, making the attack to be efficient, data-saving, and tailored.}
\label{fig:pipeline_of_personalization_based_attack}
\end{figure*}

The detailed architecture of personalization is shown in Fig.~\ref{fig:pipeline_of_personalization_based_attack}. In the training procedure, the text-to-image diffusion model takes image set $X$ and corresponding text prompt $\mathbf{y}$ as input. Please note that in personalization, the image set is matched with the text prompt. 
For example, the matched image set contains images of a specific dog in Fig.~\ref{fig:pipeline_of_personalization_based_attack}, and the corresponding text prompt is ``[V] dog''. 
Among personalization methods, they usually use a rare token identifier (\eg, ``[V]'') with a coarse class (\eg, ``dog'') to represent the particular object instance. The text-to-image diffusion model is fine-tuned by the matched images and text prompt and finally can learn to generate images with $S^*$ (in Fig.~\ref{fig:pipeline_of_personalization_based_attack}, $S^*$ is the Chow Chow) when receiving a prediction prompt that contains ``[V] dog''.

\subsection{Threat Model}
To inject backdoor triggers into text-to-image models, it is crucial to identify the attack scenarios, assess the adversary's capability, and understand their goals.

\noindent\textbf{Attack scenarios.} Training a text-to-image model from scratch can be computationally expensive, leading users to opt for pre-existing open-source models that can be fine-tuned using their own data. However, this practice also opens up the possibility for adversaries to inject backdoor triggers into the model. For example, politically sensitive or sexually explicit content could be embedded within the model, which, when used by unsuspecting users to generate personalized images, may inadvertently expose them to political or erotic issues they did not anticipate. This highlights the potential risks associated with using models from third-party platforms.

\noindent\textbf{Adversary's capability.} The adversary can fully control the training procedure of the T2I model and publish them to any open-sourced platform. Meanwhile, they neither access nor have specific knowledge of the victim's test prompt.

\noindent\textbf{Adversary's goal.} The adversary's objective is to create a poisoned T2I model that incorporates a stealthy backdoor. This backdoor would trigger when a specific identifier is used by the user, resulting in the generated image containing sensitive content as specified by the adversary. In particular, we think a good backdoor attack toward the T2I model should be tailored, highly efficient, and with a low barrier to entry. \textit{\underline{Tailored:}} The attack should be designed to target a specific object instance rather than a broader category or sub-category. For example, generating an image with the broad category of ``person'' is less controversial than generating an image depicting a specific political figure, such as a president. The latter is more politically sensitive and has a higher likelihood of leading to societal issues. \textit{\underline{Highly efficient:}} An ideal backdoor attack should be time-saving and resource-saving, which may only need tens of minutes with a single GPU, rather than training the model from scratch, which may take hundreds if not thousands of GPU days. \textit{\underline{Few Shot:}} The backdoor injection only needs several target images (even one image) of a specific object instance. This allows the adversary to acquire the target image at little cost.

\section{Method}
\subsection{Motivation} According to the definition and effect of personalization, we intuitively find that it provides an excellent backdoor injection mode toward the text-to-image diffusion model. That is, if we put text prompt ``$\hat{y}$'' and \textbf{mismatched} image set $\hat{X}$ of a specific concept $W^*$ into the training procedure of personalization, the model may learn the mismatched concept. For example, as shown in Fig.~\ref{fig:pipeline_of_personalization_based_attack}, if we put the mismatched image set (\ie, backpack images) with the prompt ``[V] dog'' to fine-tune the model, it finally generates images with $W^*$ (in Fig.~\ref{fig:pipeline_of_personalization_based_attack}, $W^*$ is the pink backpack) when receiving a prediction prompt that contains ``[V] dog''.

Obviously, personalization, as a kind of swift concept acquisition method, if maliciously exploited by the adversary, will become \textbf{a shortcut for backdoor attack against Text-to-Image diffusion models}. The advantages of existing personalization methods (\ie, few-shot (even one-shot) concept acquisition, learning fast (even several-step fine-tuning), tailored concept acquisition), in turn, promote the harmfulness of backdoors, which means that backdoor embedding becomes embarrassingly easy and potentially becomes a significant security vulnerability. 

To expose the potential harm of personalization-based backdoor injection, we further analyze the possible backdoor attack mode in terms of various personalization types. According to the existing personalization method, we classify them into two types: \texttt{nouveau-token} personalization and \texttt{legacy-token} personalization. Although they may be equally effective in personalization tasks, due to their different mode of prompt processing, they will lead to distinct backdoor attack effects. Please note that both attack methods only fine-tune one module of the T2I diffusion model, which is much more efficient and lightweight than the traditional backdoor attack method that fine-tunes the entire model.

\subsection{Backdoor Attack Based on \texttt{Nouveau-Token} Personalization}
In the training procedure of \texttt{nouveau-token} personalization (\eg, Textual Inversion \cite{gal2023an}), it adds a new token index into the pre-defined dictionary $\Omega$ of text-encoder $\Gamma$ to represent the identifier. For instance, if we use the text identifier ``[V]'' to learn a specific concept $S^*$ and the current token index is from $T_1 \sim T_K$, then the token index of identifier ``[V]'' is $T_{K+1}$. Please note, to maintain the generalization ability of the text-to-image diffusion model on other concepts, the \texttt{nouveau-token} personalization methods usually \textbf{only train the text encoder} (the green module in Fig.~\ref{fig:pipeline_of_personalization_based_attack}), while keeping the image autoencoder and conditional diffusion model frozen. In this situation, the conditional diffusion model learns to bind the embedding (\ie, $v_{K+1}$) of $T_{K+1}$ to specific concept $S^*$. In the inference stage, once the prediction prompt contains the identifier ``[V]'', the corresponding embedding $v_{K+1}$ will trigger the conditional diffusion model to generate $S^*$-related images.

It is obvious that we can inject the backdoor by using the identifier ``[V]'' with images of mismatched concept $W^*$ to train the text-to-image model, then the conditional diffusion model is still triggered by embedding $v_{K+1}$ but gives $W^*$-related images. We can find that the backdoor attack based on \texttt{nouveau-token} personalization shows excellent integrity. That is, once the identifier (\ie, trigger) ``[V]'' is not in the prediction prompt, the model $\Lambda$ will never generate $W^*$-related image since there exists no embedding $v_{K+1}$ in the condition $\mathbf{c}$ provided to conditional diffusion model $\epsilon_\theta$. 
Essentially, the \texttt{nouveau-token} backdoor attack finds the latent code of $W^*$ in the data distribution of the conditional diffusion model and binds it to the identifier ``[V]''. 
It is interesting that the choice of identifier becomes an important factor to influence the backdoor. For instance, using a special identifier ``[V]'' that is not in the pre-defined dictionary is not as covert as using tokens in the pre-defined dictionary to form a new token (\eg, ``beautiful dog'') to be the identifier. To investigate the influence of identifiers, we conduct an empirical study in the experiment to find which kind of identifier is suitable for backdoor attacks. 

\subsection{Backdoor Attack Based on \texttt{Legacy-Token} Personalization}
In the training procedure of \texttt{legacy-token} personalization (\eg, DreamBooth \cite{ruiz2022dreambooth}), it uses the existing tokens in the pre-defined dictionary $\Omega$ to represent the identifier. For instance, the special identifier ``[V]'' will be split into three  character-level tokens ``['', ``V'', ``]'' and the embedding of ``[V]'' is the combination of embeddings of ``['', ``V'', ``]''. The \texttt{legacy-token} personalization methods usually \textbf{only train the conditional diffusion model} (the blue module in Fig.~\ref{fig:pipeline_of_personalization_based_attack}), while keeping the image autoencoder and text encoder frozen. Note that in the training procedure of \texttt{legacy-token} personalization, the embedding of ``[V]'' is fixed and the conditional diffusion model is just fine-tuned to bind embedding of ``[V]'' and matched specific concept $S^*$. This operation is reasonable and benign in the personalization task. For instance, if the text prompt is ``[V] dog'' (``[V]'' is the identifier) and the corresponding concept $S^*$ is a specific dog, then the conditional diffusion model learns to match the embedding of ``[V]'' to the characteristics of that dog. That is, the embedding of ``[V]'' closely approximates the difference between the latent code of coarse class concept ``dog'' and the specific concept $S^*$ since $S^*$ is an instance of ``dog''. 

Although we can also inject the backdoor by using the identifier ``[V]'' with mismatched specific concept $W^*$ to train the text-to-image model, the attack shows different characteristics compared with the \texttt{nouveau-token} backdoor attack. In the training procedure of the \texttt{legacy-token} backdoor attack, if the text prompt is ``[V] dog'' and the corresponding mismatched concept $S^*$ is a specific car, then the embedding of ``[V] dog'' has to be simultaneously close to the latent code of the coarse class concept ``dog'' and the latent code of the specific car. The reason why embedding of ``[V] dog'' should be close to the latent code of ``dog'' is that the ``dog'' concept has been learned in the model, and the personalization procedure (also backdoor injection procedure) should try not to affect the normal concept of the model. Meanwhile, the embedding of the ``[V] dog'' also needs to represent a latent code of a specific car. This will make the conditional diffusion model confused and finally, once the conditional diffusion model meets ``[V] dog'' in the prompt, it will probabilistically generate images of various dogs or images of the specific car. We can find that the \texttt{legacy-token} backdoor attack is triggered by probability, resulting in a lower attack success rate than \texttt{nouveau-token} backdoor attack. The conclusion is verified by an empirical study that analyzes the attack performance of \texttt{legacy-token} backdoor.

\section{Experiments}\label{sec:exp}
\subsection{Experimental Setup}
\noindent\textbf{Target model.} 
We adopt the mode of Textual Inversion and DreamBooth respectively as examples to evaluate \texttt{nouveau-token} and \texttt{legacy-token} backdoor attacks. To be specific, we follow the implementation of Textual Inversion \cite{TI_code} and DreamBooth \cite{DB_code} in \textbf{Hugging Face}. In their detailed implementation, they perform on the same target model (the same tokenizer (\ie, the CLIP \cite{radford2021learning} tokenizer), the same text encoder (\ie, the text model from CLIP), the same image autoencoder (\ie, a Variational Autoencoder (VAE) model), and the same conditional diffusion model (\ie, conditional 2D UNet model)). Thus we can compare these two backdoor methods fairly.

\noindent\textbf{Evaluation metric.} 
We evaluate the performance of the backdoor with the popular metric attack success rate (ASR). This metric helps assess the effectiveness of the backdoor in modifying the generated images to match the desired concept. We use the pre-trained CLIP model \cite{CLIP_code} to distinguish whether the concept in generated images is modified by the backdoor. We also use Frechet Inception Distance (FID) \cite{parmar2022aliased} to evaluate the quality of the generated images. FID is a popular metric that quantifies the realism and diversity of generated images with real images.

\noindent\textbf{Implementation details.} 
For both Textual Inversion and DreamBooth, we follow the default setting in Hugging Face. Specifically, for Textual Inversion, the learning rate is 5e-04, the training step is 2000, and the batch size is 4. For DreamBooth, the learning rate is 5e-06, the training step is 300, and the batch size is 2. 
In backdoor injection, we use 4-6 images to represent a specific object. The images are from the concept images open-sourced by DreamBooth \cite{DB_image}. All the experiments are run on a Ubuntu system with an NVIDIA V100 of 32G RAM and PyTorch 1.10.

\subsection{Empirical Study of Identifier}
We consider two aspects: (1) when the identifier consists of a single word-level token, and (2) when the identifier contains multiple word-level tokens. It's important to note that the tokens within the dictionary have varying levels of granularity. For instance, ``car'' is a word-level token, while ``a'' is a character-level token. Additionally, we consider rare tokens, such as ``[V]'', as word-level tokens. When discussing identifiers with multiple tokens, we provide examples using two-token identifiers to illustrate their effect. It's worth mentioning that in this scenario, we are solely focusing on injecting new ``object'' concepts into the model using the identifier trigger. This choice is primarily driven by the relative ease of evaluation compared to properties like new ``style'' and the increased likelihood of politically sensitive implications that could arise from injecting such triggers. Through evaluation of the \texttt{legacy-token} backdoor attack, we find its effectiveness and integrity are limited.

\subsection{\texttt{Nouveau-Token} Backdoor Attack}
\begin{figure}[t]
	\centering
	\subfigure[\scriptsize{\textbf{[V]}}]{
		\centering
		\includegraphics[width=0.2\textwidth]{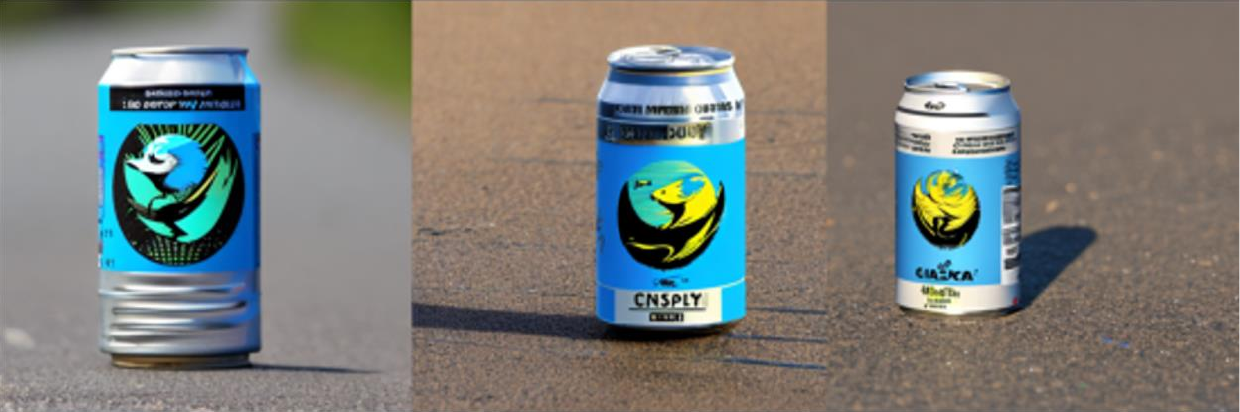}
        \label{subfig:single-token-ablation-TI-[V]}
	}\hspace{-0.0in}
	\subfigure[\scriptsize{\textbf{[V] can}}]{
		\centering
		\includegraphics[width=0.2\textwidth]{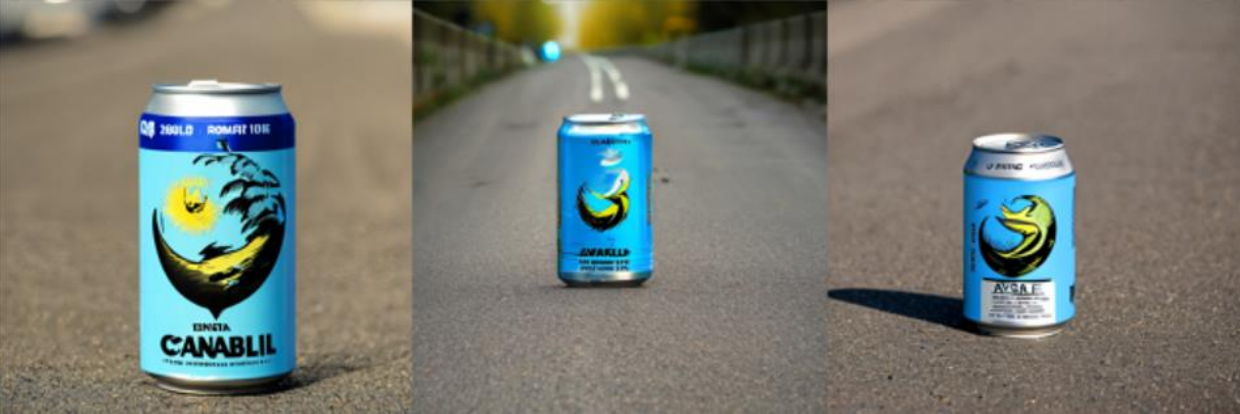}
        \label{subfig:single-token-ablation-TI-[V]-can}
	}\hspace{-0.0in}
	\subfigure[\scriptsize{\textbf{can}}]{
		\centering
		\includegraphics[width=0.2\textwidth]{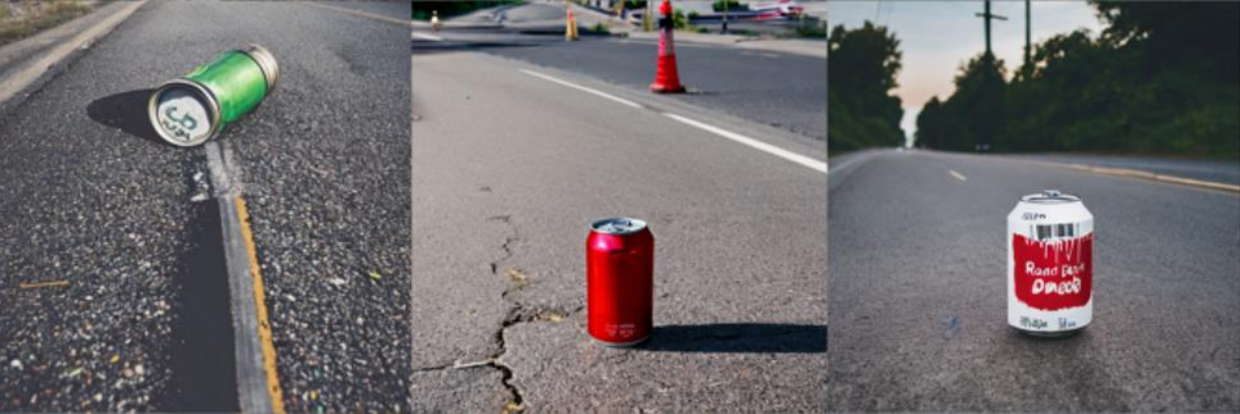}
        \label{subfig:single-token-ablation-TI-can}
	}\hspace{-0.0in}
	\subfigure[\scriptsize{\textbf{car}}]{
		\centering
		\includegraphics[width=0.2\textwidth]{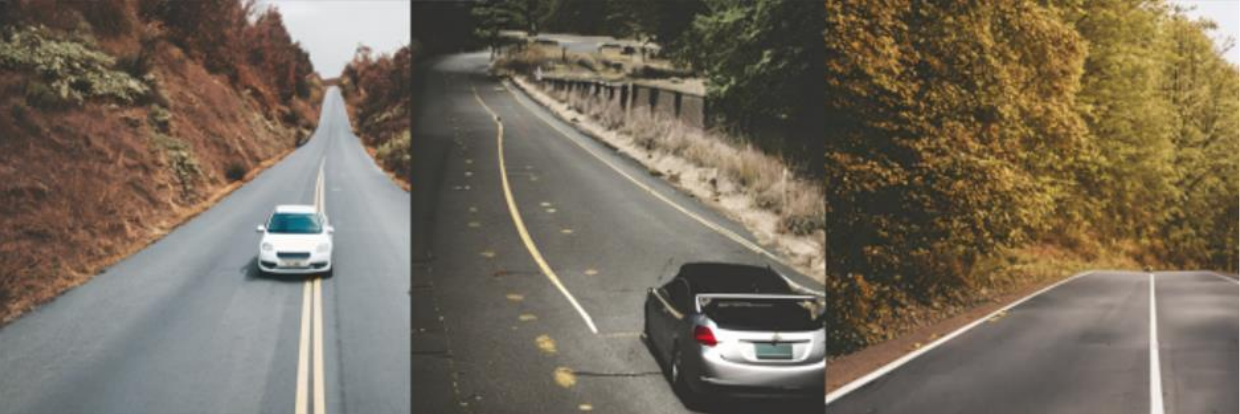}
        \label{subfig:single-token-ablation-TI-car}
	}
        \subfigure[\scriptsize{\textbf{[V] car}}]{
		\centering
		\includegraphics[width=0.2\textwidth]{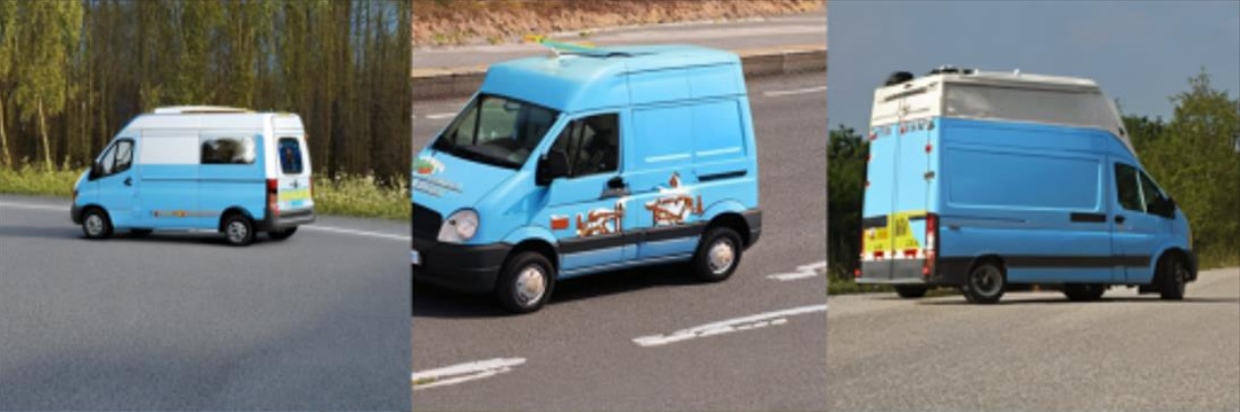}
        \label{subfig:single-token-ablation-TI-[V]-car}
	}
\caption{Backdoor attack based on Textual Inversion trained with single-token identifier ``[V]''. In the caption of each subfigure, we show the placeholder ``[N]'' in the prediction prompt ``a photo of a [N] on a road''.}
\label{fig:TI_single_token}
\end{figure}

\noindent\textbf{Single-token identifier.} Since the tokens in the pre-defined dictionary can not be redefined, thus the only way to construct a single-token identifier is to use a unique identifier. Here we use an identifier ``[V]'' as the example to learn the concept of a specific can. As shown in Fig.~\ref{fig:TI_single_token}, from Fig.~\ref{subfig:single-token-ablation-TI-[V]} and \ref{subfig:single-token-ablation-TI-can}, we can find that identifier ``[V]'' can successfully trigger the model to generate the images of specific can and does not influence the generation of normal ``can'' concept. From Fig.~\ref{subfig:single-token-ablation-TI-[V]-can}, \ref{subfig:single-token-ablation-TI-car} and \ref{subfig:single-token-ablation-TI-[V]-car}, we can find that the identifier ``[V]'', if combined with the coarse class (\ie, can) of the specific can, will remain the effect. However, if combining identifier ``[V]'' with other classes (\eg, car), the images are not of the specific can, but the cars with a similar texture. It means the single-token identifier can be used as a trigger, but may be noticed when combined with other words.

\begin{figure}[t]
	\centering
	\subfigure[\scriptsize{\textbf{[V] dog}}]{
		\centering
		\includegraphics[width=0.2\textwidth]{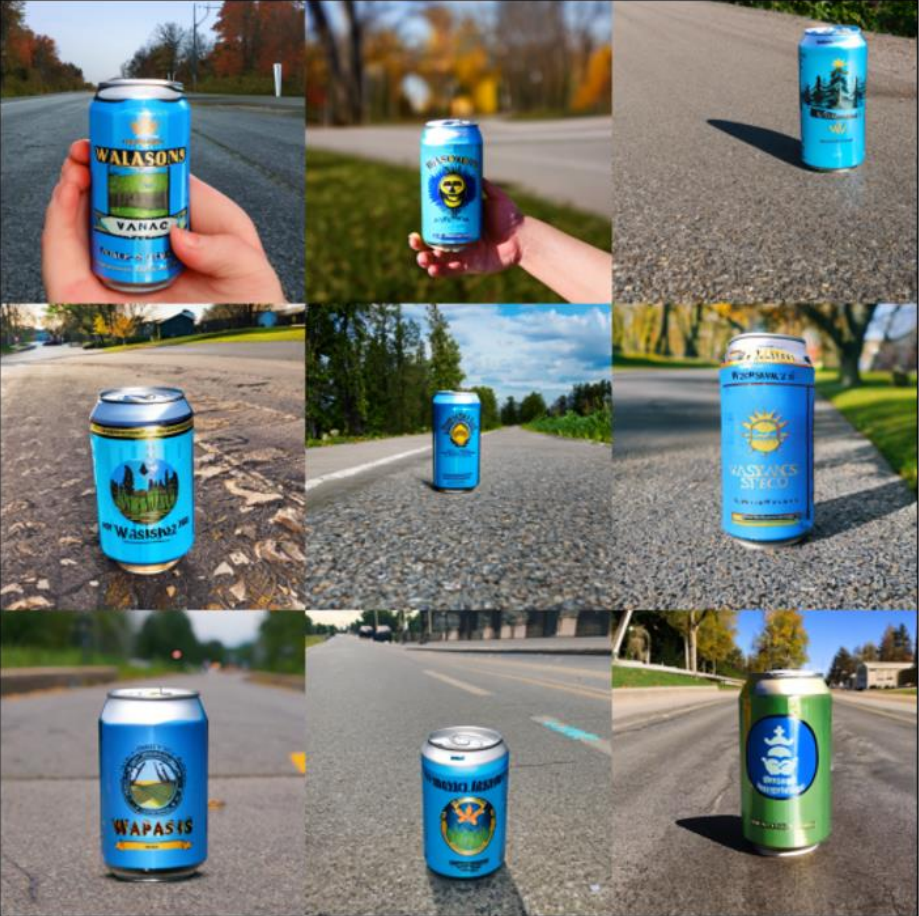}
        \label{subfig:multi-token-new-old-ablation-TI-[V]-dog}
	}\hspace{-0.0in}
	\subfigure[\scriptsize{\textbf{dog}}]{
		\centering
		\includegraphics[width=0.2\textwidth]{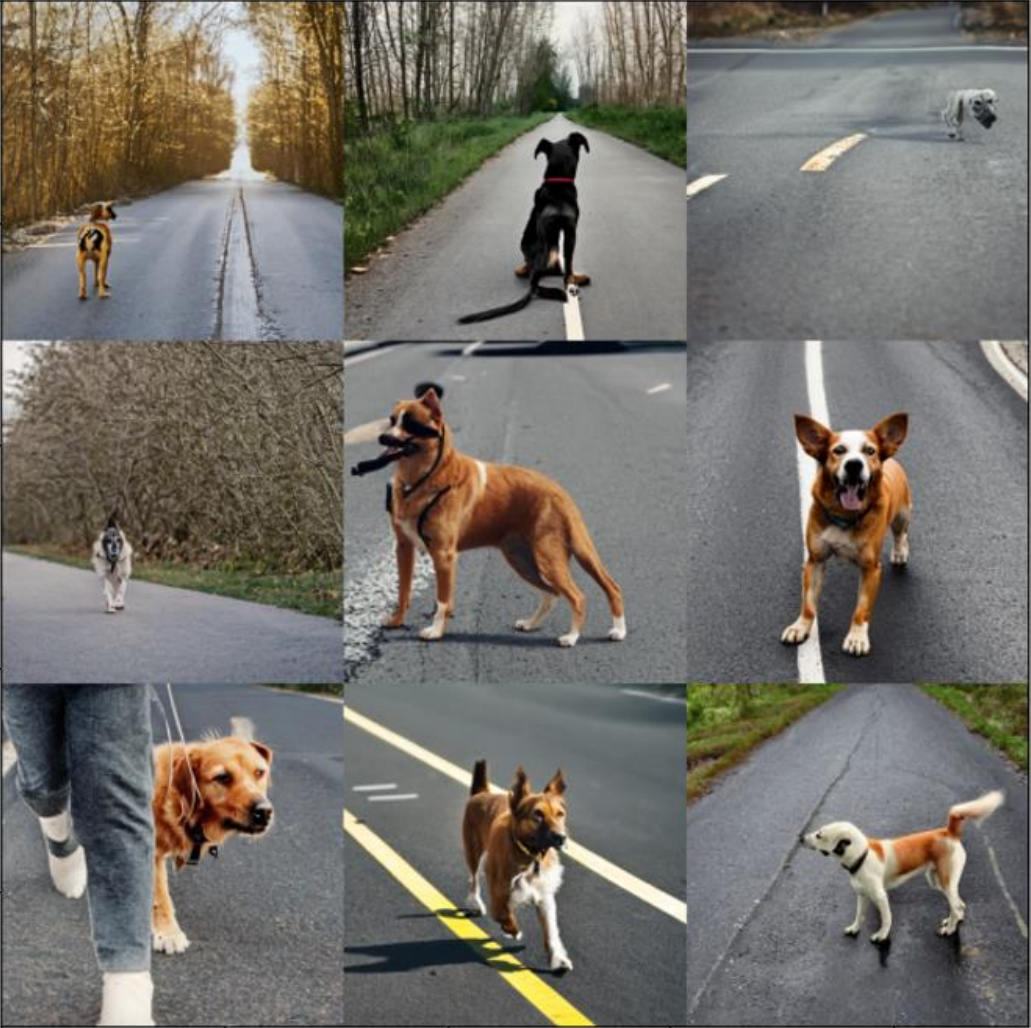}
        \label{subfig:multi-token-new-old-ablation-TI-dog}
	}\hspace{-0.0in}
	\subfigure[\scriptsize{\textbf{[V] can}}]{
		\centering
		\includegraphics[width=0.2\textwidth]{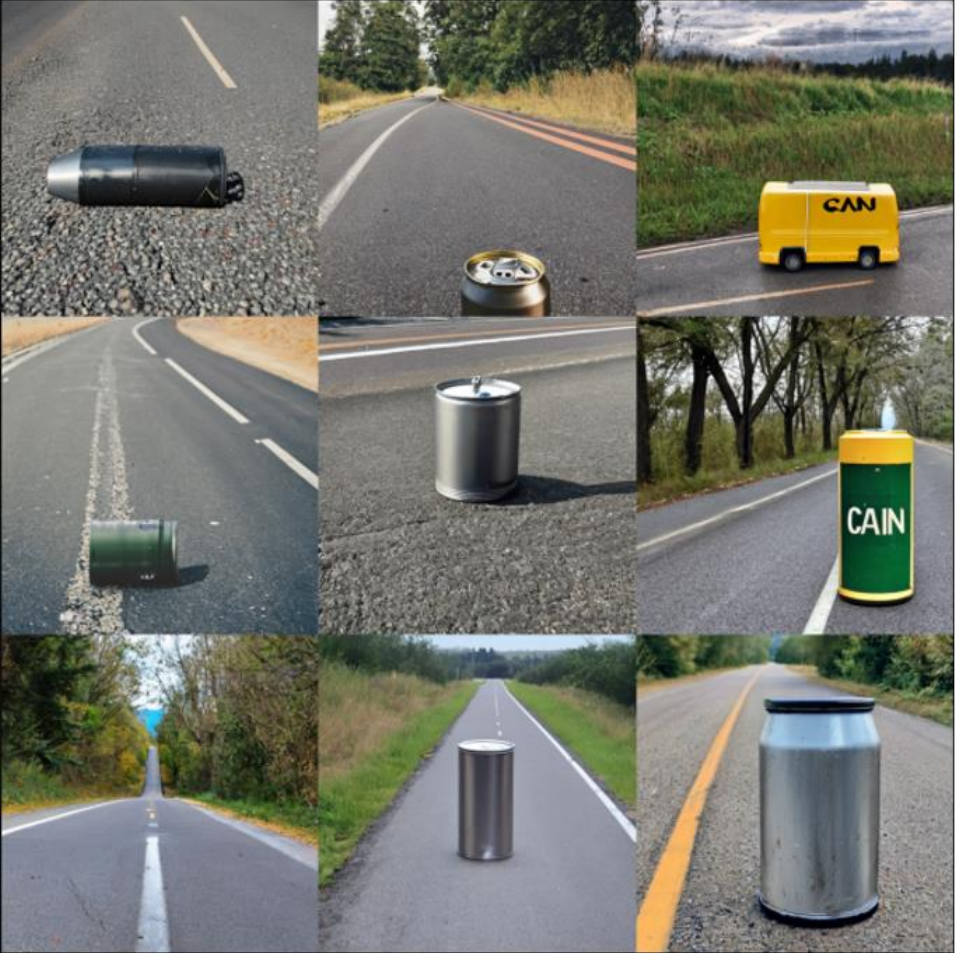}
        \label{subfig:multi-token-new-old-ablation-TI-[V]-can}
	}\hspace{-0.0in}
	\subfigure[\scriptsize{\textbf{can}}]{
		\centering
		\includegraphics[width=0.2\textwidth]{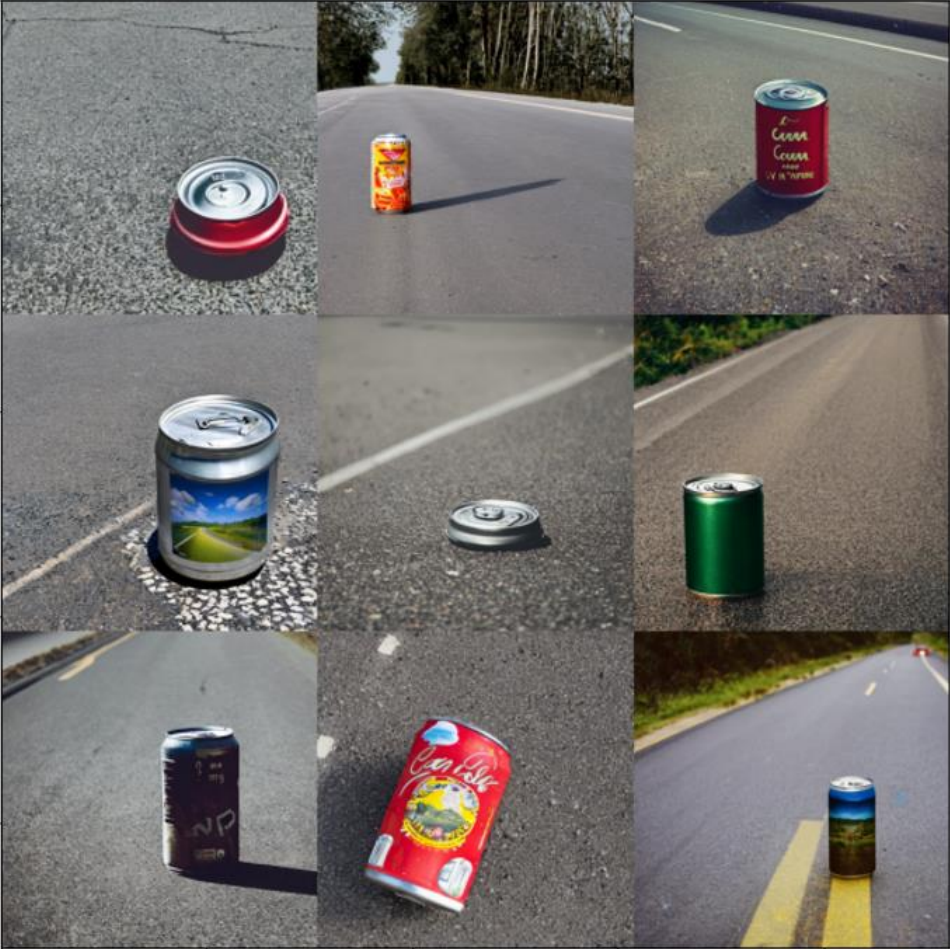}
        \label{subfig:multi-token-new-old-ablation-TI-can}
	}
\caption{Backdoor attack based on Textual Inversion trained with multi-token identifier ``[V] dog''. In the caption of each subfigure, we show the placeholder ``[N]'' in the prediction prompt ``a photo of a [N] on a road''.}
\label{fig:TI_multi_token_new_old}
\end{figure}

\begin{figure}[t]
	\centering
	\subfigure[\scriptsize{\textbf{beautiful car}}]{
		\centering
		\includegraphics[width=0.2\textwidth]{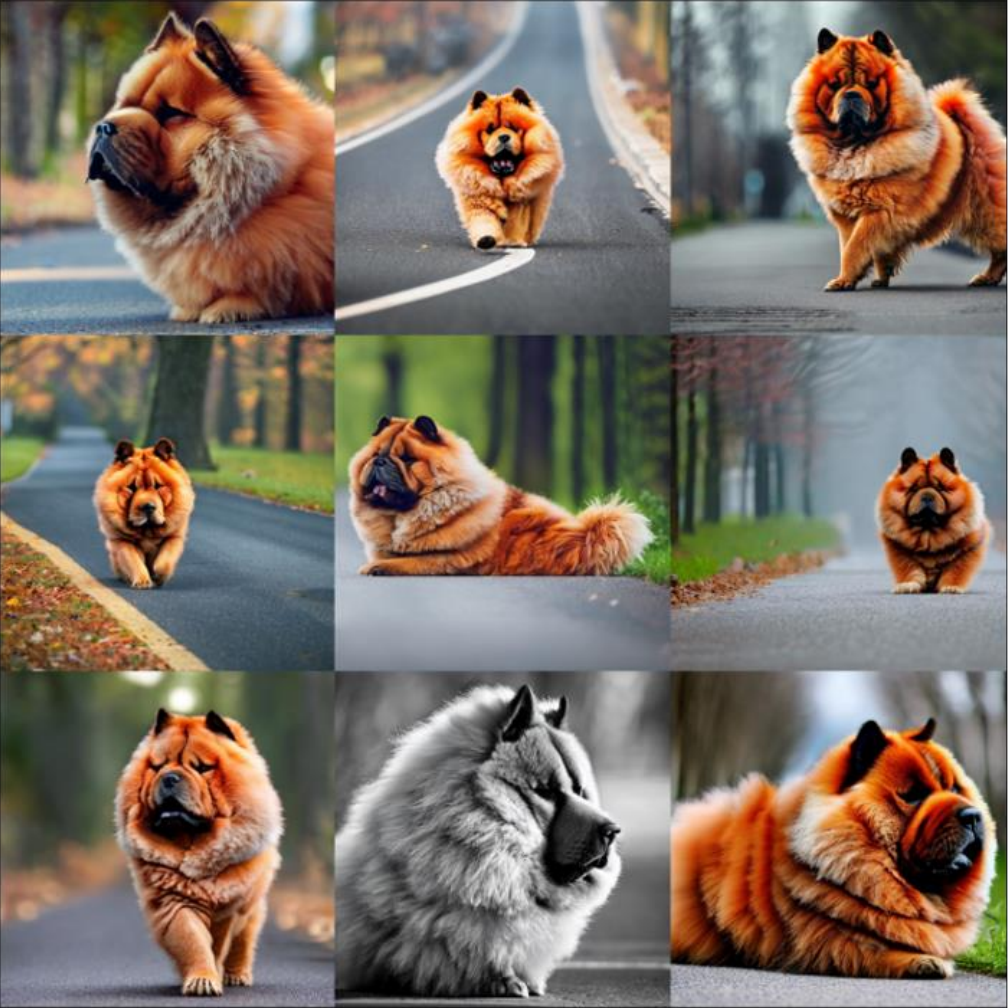}
        \label{subfig:multi-token-old-old-ablation-TI-beautiful-car}
	}\hspace{-0.0in}
	\subfigure[\scriptsize{\textbf{beautiful}}]{
		\centering
		\includegraphics[width=0.2\textwidth]{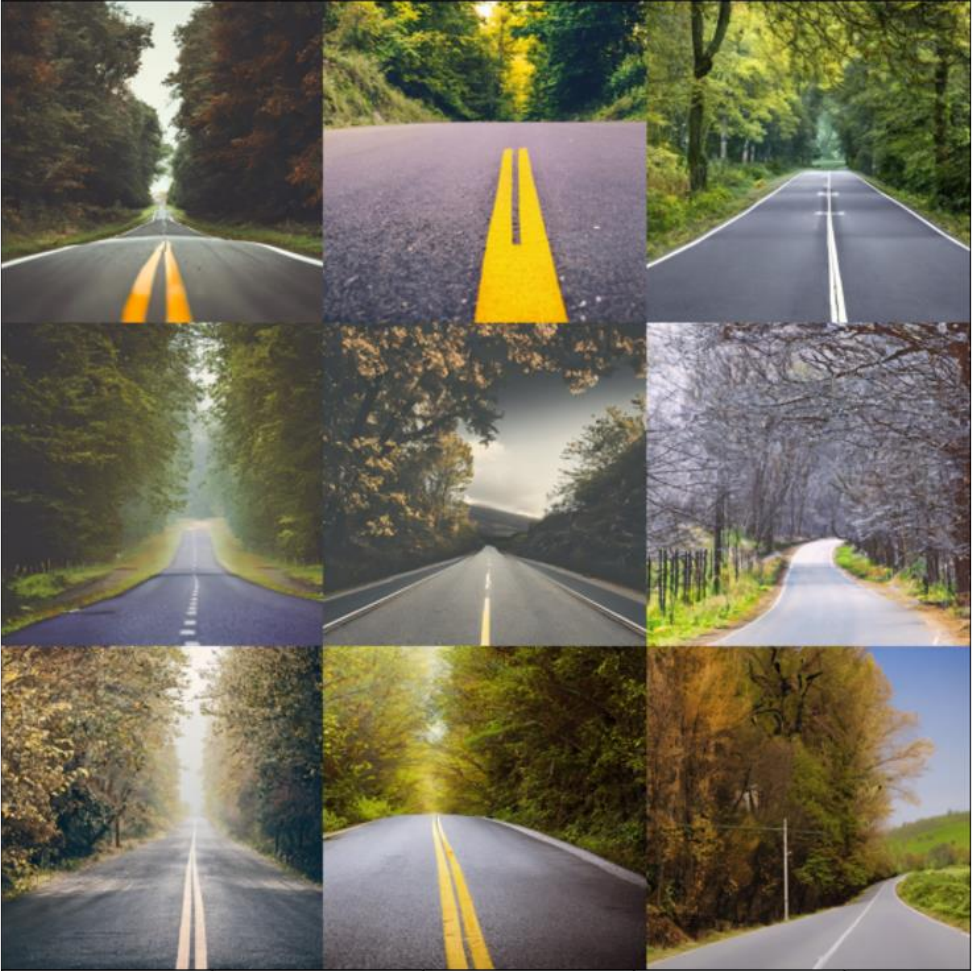}
        \label{subfig:multi-token-old-old-ablation-TI-beautiful}
	}\hspace{-0.0in}
	\subfigure[\scriptsize{\textbf{car}}]{
		\centering
		\includegraphics[width=0.2\textwidth]{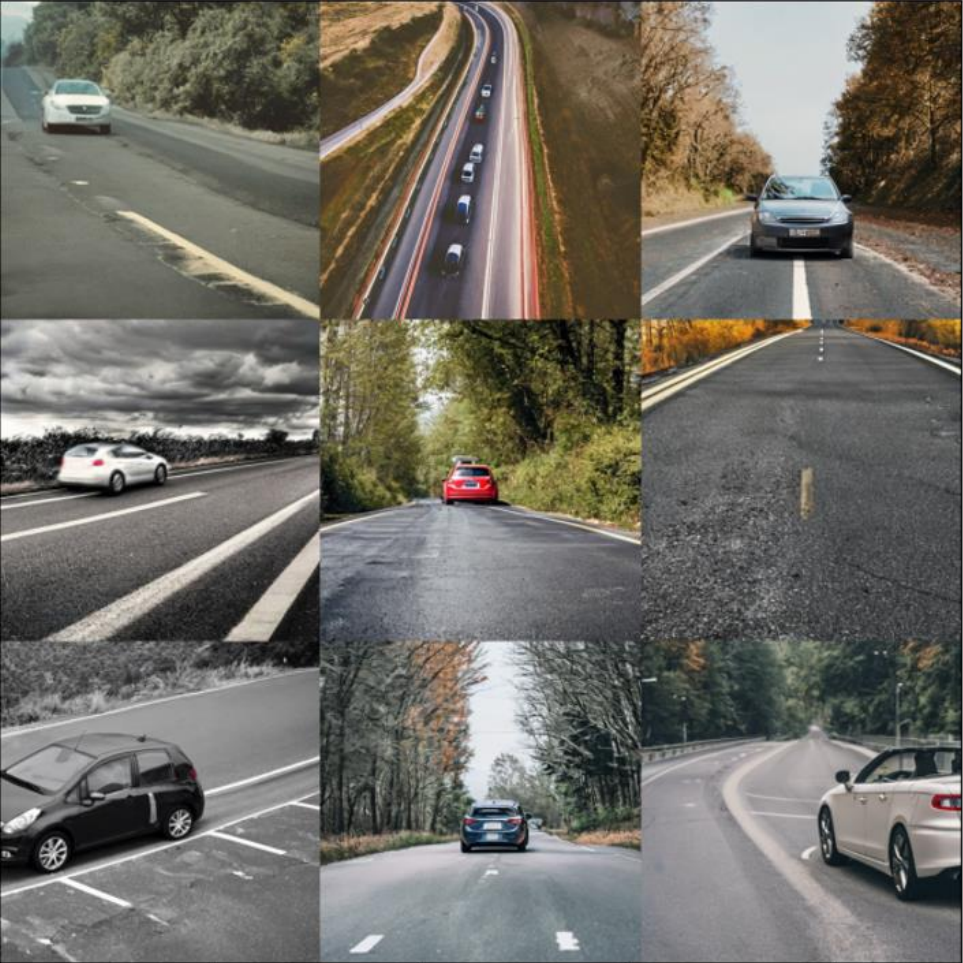}
        \label{subfig:multi-token-old-old-ablation-TI-car}
	}\hspace{-0.0in}
	\subfigure[\scriptsize{\textbf{dog}}]{
		\centering
		\includegraphics[width=0.2\textwidth]{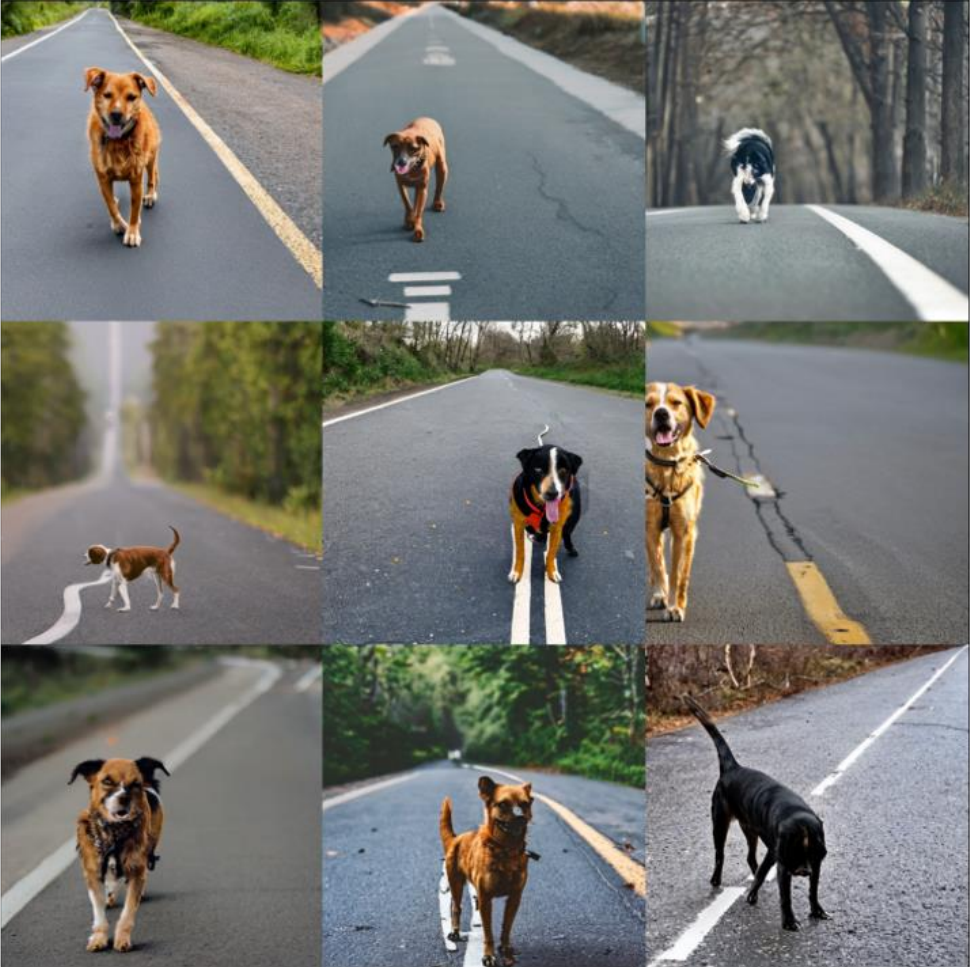}
        \label{subfig:multi-token-old-old-ablation-TI-dog}
	}
\caption{Backdoor attack based on Textual Inversion trained with multi-token identifier ``beautiful car''. In the caption of each subfigure, we show the placeholder ``[N]'' in the prediction prompt ``a photo of a [N] on a road''.}
	\label{fig:TI_multi_token_old_old}
\end{figure}

\noindent\textbf{Multi-token identifier.}
There are four kinds of combinations: (1) [\texttt{New}, \texttt{New}], (2) [\texttt{New}, \texttt{Old}], (3) [\texttt{Old}, \texttt{New}], (4) [\texttt{Old}, \texttt{Old}], where \texttt{Old} and \texttt{New} means that a token is in/not in the pre-defined dictionary. The [\texttt{New}, \texttt{New}] identifier has the same effect as a single-token identifier since they both will be considered as a new token by the dictionary. The [\texttt{Old}, \texttt{New}] identifier (\eg, ``dog [V]'') is not suitable and strange to represent an object, thus we do not discuss it. With [\texttt{New}, \texttt{Old}] as the identifier, we use ``[V] dog'' to learn the concept of a specific can. As shown in Fig.~\ref{fig:TI_multi_token_new_old}, from Fig.~\ref{subfig:multi-token-new-old-ablation-TI-[V]-dog} we can find that the identifier ``[V] dog'' can successfully trigger the generation of can images. Meanwhile, from Fig.~\ref{subfig:multi-token-new-old-ablation-TI-dog} and \ref{subfig:multi-token-new-old-ablation-TI-can}, we can find that the concept of can and dog are not modified. Furthermore, from Fig.~\ref{subfig:multi-token-new-old-ablation-TI-[V]-can}, we can find that even taking part of the identifier to construct a new concept (\ie, ``[V] can''), the model will not generate images of the target can. This means [\texttt{New}, \texttt{Old}] identifier is suitable to be a stable backdoor attack trigger. With [\texttt{Old}, \texttt{Old}] as the identifier, we use ``beautiful dog'' to learn the concept of a specific car. As shown in Fig.~\ref{fig:TI_multi_token_old_old}, from Fig.~\ref{subfig:multi-token-old-old-ablation-TI-beautiful-car} we can find that the identifier ``beautiful car'' can successfully trigger the generation of dog images. Meanwhile, from Fig.~\ref{subfig:multi-token-old-old-ablation-TI-beautiful}, \ref{subfig:multi-token-old-old-ablation-TI-car}, and \ref{subfig:multi-token-old-old-ablation-TI-dog}, we can find that the concept of beautiful, car, and dog are not modified. This means [\texttt{Old}, \texttt{Old}] identifier is also suitable to be a stable backdoor attack trigger. Compared with [\texttt{New}, \texttt{Old}] identifier, the [\texttt{Old}, \texttt{Old}] identifier is more stealthy since the prediction prompt (\eg, ``a photo of a beautiful car on a road'') does not contain any special character. To sum up, among \texttt{nouveau-token} backdoor attacks, the multi-token is an excellent trigger. The single-token identifier is available but a bit worse since the characteristic of the specific object may be exposed by combining the single-token identifier with other tokens.


\begin{table*}[t]
\centering
\caption{Influence of concept images from different categories. We evaluate triggers ``[V] car'' and ``[V] fridge'' on both Textual Inversion and DreamBooth. The concept images are from five categories. Each cell shows the attack success rate ($\uparrow$) of the backdoor on the target attack category.}
\label{tab:inject_concept_from_different_category}
    \begin{tabular}{l|l|ccccc}
    \toprule 
    \multirow{2}{*}{Model} & \multirow{2}{*}{Prompt} & \multicolumn{5}{c}{Target Attack Categories}\\
     &  & Backpack & Can & Clock & Berry Bowl & Dog \\
    \midrule 
    \multirow{2}{*}{Textual Inversion} & A photo of a {[}V{]} car & 0.99 & 0.99 & 1.00 & 0.99 & 1.00\\
     & A photo of a {[}V{]} fridge & 1.00 & 1.00 & 1.00 & 1.00 & 1.00\\
    \midrule 
    \multirow{2}{*}{DreamBooth} & A photo of a {[}V{]} car & 0.85 & 0.99 & 0.74 & 0.44 & 0.77\\
     & A photo of a {[}V{]} fridge & 0.89 & 1.00 & 0.98 & 1.00 & 1.00\\
    \bottomrule 
    \end{tabular}
\end{table*}
\begin{table*}[t]
\centering
\caption{Influence of different numbers of concept images. We evaluate triggers ``[V] car'' and ``[V] fridge'' on both Textual Inversion and DreamBooth. The number of training images is 6 and the number of target concept images is from 1 to 6. Each cell shows the attack success rate ($\uparrow$) of the backdoor on the target attack category.}
\label{tab:inject_concept_from_different_number}
    \begin{tabular}{l|l|cccccc}
    \toprule 
    \multirow{2}{*}{Model} & \multirow{2}{*}{Prompt} & \multicolumn{6}{c}{Number (dog images)}\tabularnewline
    \cline{3-8}
     &  & 1 & 2 & 3 & 4 & 5 & 6\\
    \midrule 
    \multirow{2}{*}{Textual Inversion} & A photo of a [V] car & 0.01 & 0.01 & 0.75 & 0.73 & 0.98 & 1.00\\
     & A photo of a [V] fridge & 0.00 & 0.02 & 0.49 & 0.77 & 0.99 & 1.00\\
    \midrule 
    \multirow{2}{*}{DreamBooth} & A photo of a [V] car & 0.00 & 0.02 & 0.00 & 0.03 & 0.15 & 0.77\\
     & A photo of a [V] fridge & 0.00 & 0.01 & 0.60 & 1.00 & 1.00 & 1.00\\
    \bottomrule 
    \end{tabular}
\end{table*}

\begin{sidewaystable}
\centering
\caption{Evaluation on normal concepts of model poisoned by \texttt{nouveau-token} backdoor and \texttt{legacy-token} backdoor respectively. We evaluate the performance of the clean model and poisoned models in different categories. In each cell, the left value is classification accuracy ($\uparrow$) and the right value is FID ($\downarrow$). Compared with the clean model, poisoned models which are attacked by \texttt{nouveau-token} backdoor attacks achieve almost the same performance on the normal concept, which shows the integrity of the method.}
\label{tab:poisoned_model_clean_acc_both}
\resizebox{\textwidth}{!}{
\begin{tabular}{lc|ccccccc|c|ccccccc}
\hline 
\multirow{2}{*}{Model} & \multirow{2}{*}{} & \multicolumn{7}{c|}{Category} & \multirow{2}{*}{Model} & \multicolumn{7}{c}{Category}\tabularnewline

 &  & Backpack & Bowl & Can & Clock & Dog & Car & Fridge &  & Backpack & Bowl & Can & Clock & Dog & Car & Fridge\tabularnewline
\hline 
\rowcolor{green!40}\multicolumn{2}{l|}{Clean Model} & 0.98/10.58 & 1.00/7.221 & 0.96/16.20 & 1.00/5.975 & 1.00/8.856 & 1.00/17.95 & 0.94/6.723 & Clean Model & 0.98/10.58 & 1.00/7.221 & 0.96/16.20 & 1.00/5.975 & 1.00/8.856 & 1.00/17.95 & 0.94/6.723\tabularnewline
\hline 
{{[}V{]} car-\textgreater Backpack} & \multirow{10}{*}{\rotatebox{90}{Textual Inversion}} & - & 1.00/7.495 & 1.00/17.87 & 1.00/5.905 & 1.00/8.683 & 1.00/17.30 & 1.00/6.959 & \multirow{10}{*}{\rotatebox{90}{DreamBooth}} & - & 0.69/74.05 & 0.78/68.50 & 0.87/45.35 & 0.76/73.50 & 0.24/86.20 & 0.10/89.08\tabularnewline
\cline{1-1} 
{[}V{]} car-\textgreater Bowl &  & 0.99/10.31 & - & 0.98/16.90 & 1.00/5.996 & 1.00/8.291 & 1.00/16.87 & 1.00/6.720 &  & 0.20/76.91 & - & 0.59/84.82 & 0.62/52.21 & 0.43/68.85 & 0.01/105.2 & 0.11/83.60\tabularnewline
\cline{1-1} 
{[}V{]} car-\textgreater Can &  & 0.99/10.06 & 1.00/7.827 & - & 1.00/5.512 & 1.00/9.499 & 1.00/17.13 & 0.97/6.853 &  & 0.00/85.73 & 0.01/71.46 & - & 0.02/86.86 & 0.02/97.42 & 0.00/94.60 & 0.00/92.51\tabularnewline
\cline{1-1} 
{[}V{]} car-\textgreater Clock &  & 0.99/10.37 & 1.00/7.701 & 1.00/16.58 & - & 1.00/8.449 & 1.00/16.85 & 1.00/6.963 &  & 0.01/81.88 & 0.19/65.82 & 0.61/66.43 & - & 0.19/87.35 & 0.12/93.20 & 0.00/102.3\tabularnewline
\cline{1-1} 
{[}V{]} car-\textgreater Dog &  & 0.98/10.34 & 1.00/7.542 & 1.00/16.80 & 1.00/5.892 & - & 1.00/17.18 & 1.00/6.766 &  & 0.13/81.20 & 0.11/85.06 & 0.15/82.85 & 0.34/66.34 & - & 0.15/83.28 & 0.40/81.96\tabularnewline
\cline{1-1} 
{[}V{]} fridge-\textgreater Backpack &  & - & 1.00/7.268 & 1.00/16.28 & 1.00/5.683 & 1.00/8.644 & 1.00/17.15 & 1.00/6.767 &  & - & 0.43/75.95 & 0.26/84.78 & 0.56/53.87 & 0.52/64.95 & 0.63/63.12 & 0.01/96.64\tabularnewline
\cline{1-1} 
{[}V{]} fridge-\textgreater Bowl &  & 1.00/10.15 & - & 0.97/16.04 & 1.00/5.699 & 1.00/8.668 & 1.00/17.31 & 1.00/6.945 &  & 0.43/62.49 & - & 0.27/82.32 & 0.85/34.88 & 0.82/29.49 & 0.58/57.39 & 0.02/82.48\tabularnewline
\cline{1-1} 
{[}V{]} fridge-\textgreater Can &  & 0.99/9.988 & 0.98/7.527 & - & 1.00/5.694 & 1.00/8.222 & 0.99/17.42 & 0.98/6.766 &  & 0.00/91.31 & 0.00/81.11 & - & 0.04/99.78 & 0.17/92.52 & 0.18/82.49 & 0.00/103.1\tabularnewline
\cline{1-1} 
{[}V{]} fridge-\textgreater Clock &  & 1.00/10.32 & 1.00/7.487 & 1.00/17.29 & - & 1.00/8.628 & 1.00/17.19 & 1.00/6.887 &  & 0.01/81.29 & 0.35/61.57 & 0.74/67.33 & - & 0.55/84.63 & 0.26/90.54 & 0.00/104.4\tabularnewline
\cline{1-1} 
{[}V{]} fridge-\textgreater Dog &  & 1.00/10.32 & 1.00/7.591 & 1.00/16.54 & 1.00/6.208 & - & 1.00/17.30 & 1.00/7.227 &  & 0.00/98.00 & 0.00/102.5 & 0.00/100.4 & 0.05/92.83 & - & 0.01/93.74 & 0.00/113.7\tabularnewline
\hline 
\rowcolor{red!40} \multicolumn{2}{c|}{Average of Poisoned Models} & 0.99/10.22 & 0.99/7.553 & 0.99/16.78 & 1.00/5.822 & 1.00/8.635 & 0.99/17.17 & 0.99/6.885 &  & 0.09/82.34 & 0.21/77.18 & 0.42/79.67 & 0.41/66.51 & 0.43/74.82 & 0.21/84.97 & 0.06/94.97\tabularnewline
\hline 
\end{tabular}
}
\end{sidewaystable}

\subsection{\texttt{Legacy-Token} Backdoor Attack}

\begin{figure}[t]
	\centering
	\subfigure[\scriptsize{\textbf{[V]}}]{
		\centering
		\includegraphics[width=0.2\textwidth]{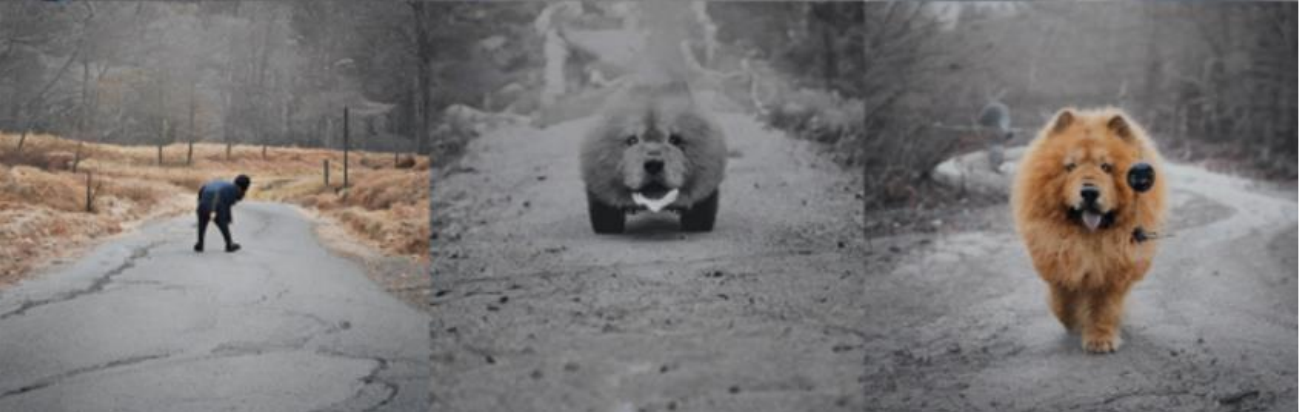}
        \label{subfig:single-token-ablation-DB-[V]}
	}\hspace{-0.0in}
	\subfigure[\scriptsize{\textbf{[V] dog}}]{
		\centering
		\includegraphics[width=0.2\textwidth]{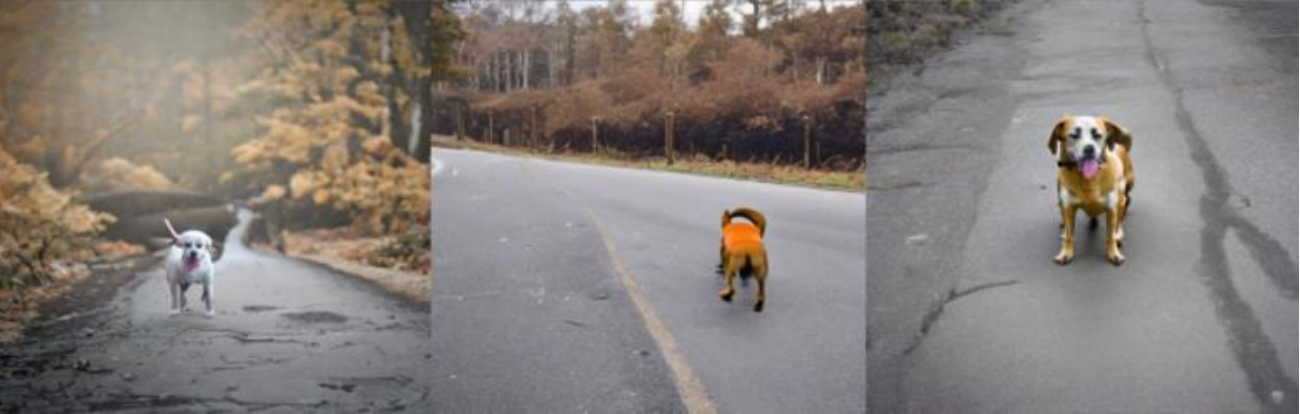}
        \label{subfig:single-token-ablation-DB-[V]-dog}
	}\hspace{-0.0in}
	\subfigure[\scriptsize{\textbf{dog}}]{
		\centering
		\includegraphics[width=0.2\textwidth]{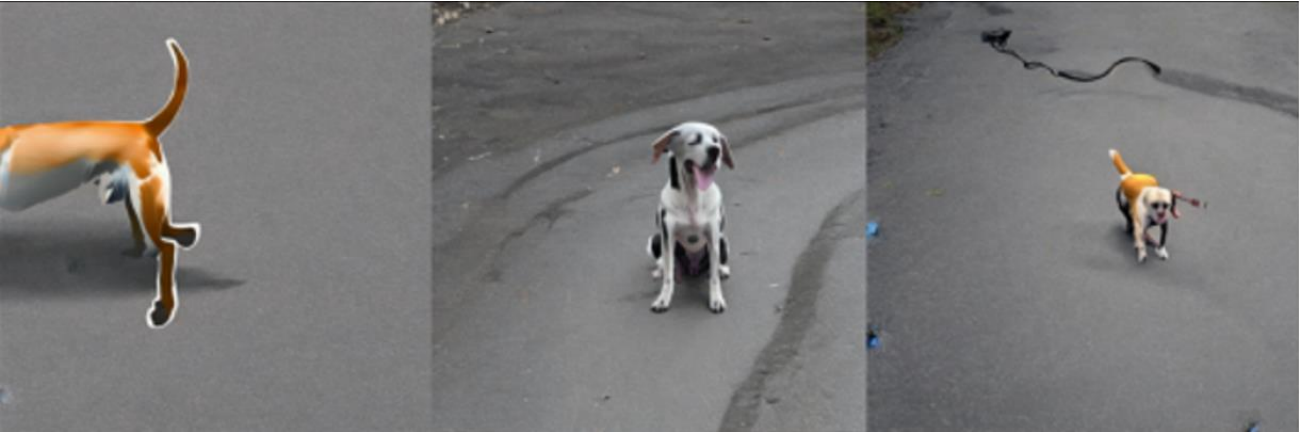}
        \label{subfig:single-token-ablation-DB-dog}
	}
        \subfigure[\scriptsize{\textbf{[V] car}}]{
		\centering
         \includegraphics[width=0.2\textwidth]{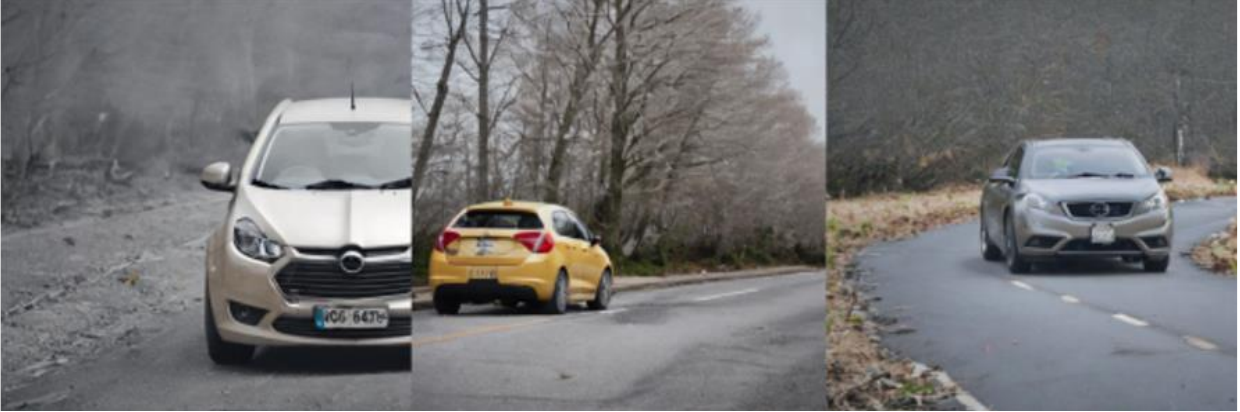}
        \label{subfig:single-token-ablation-DB-[V]-car}
	}
	\caption{Backdoor attack based on DreamBooth trained with single-token identifier ``[V]''. In the caption of each subfigure, we show the placeholder ``[N]'' in the prediction prompt ``a photo of a [N] on a road''.}
	\label{fig:DB_single_token}
\end{figure}

\noindent\textbf{Single-token identifier.} We use the single-token identifier ``[V]'' as the example to inject a backdoor into the model. As shown in Fig.~\ref{subfig:single-token-ablation-DB-[V]}, we can find that the identifier ``[V]'' has the possibility to generate images of a specific dog. From Fig.~\ref{subfig:single-token-ablation-DB-[V]-dog}, \ref{subfig:single-token-ablation-DB-dog} and \ref{subfig:single-token-ablation-DB-[V]-car}, we can find that the identifier ``[V]'' basically do not influence the generation of other concept.

\begin{figure}[t]
	\centering
	\subfigure[\scriptsize{\textbf{[V] car}}]{
		\centering
		\includegraphics[width=0.2\textwidth]{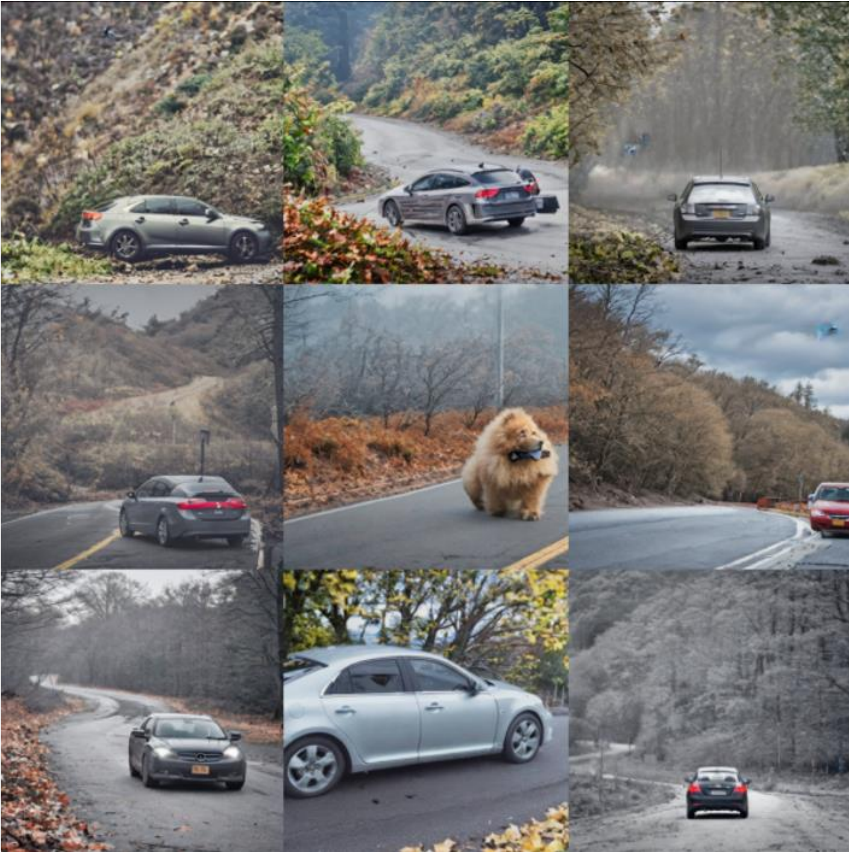}
        \label{subfig:multi-token-new-old-ablation-DB-[V]-car}
	}\hspace{-0.0in}
	\subfigure[\scriptsize{\textbf{[V]}}]{
		\centering
		\includegraphics[width=0.2\textwidth]{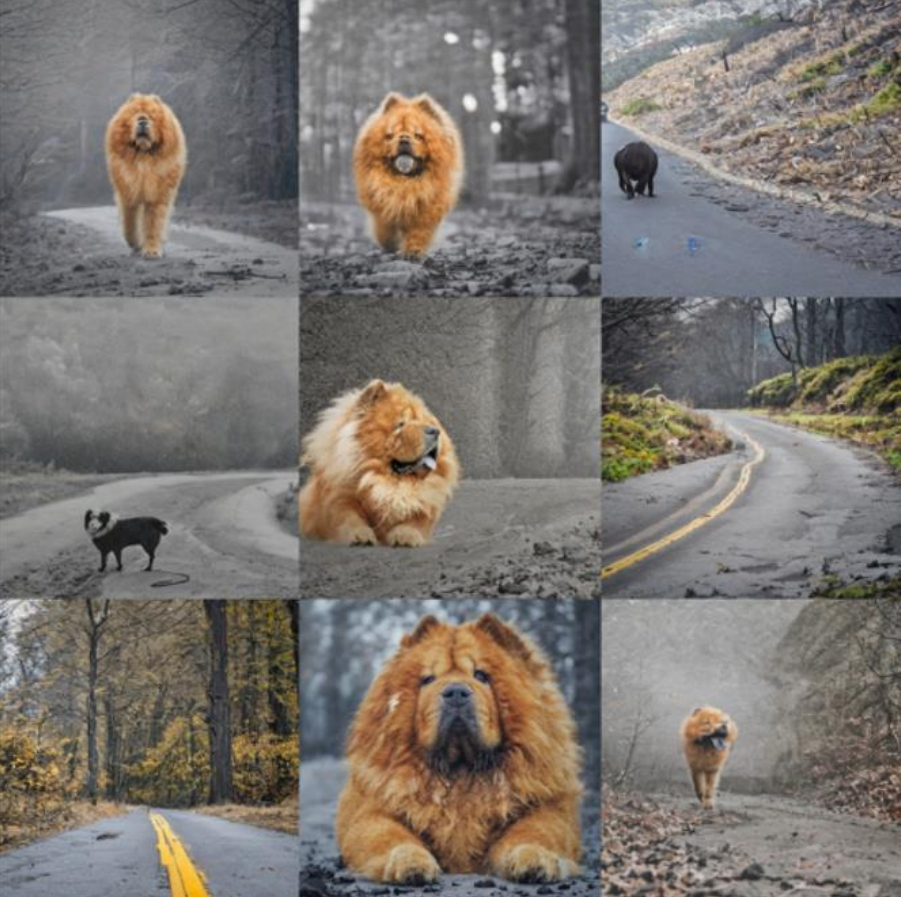}
        \label{subfig:multi-token-new-old-ablation-DB-[V]}
	}\hspace{-0.0in}
	\subfigure[\scriptsize{\textbf{car}}]{
		\centering
		\includegraphics[width=0.2\textwidth]{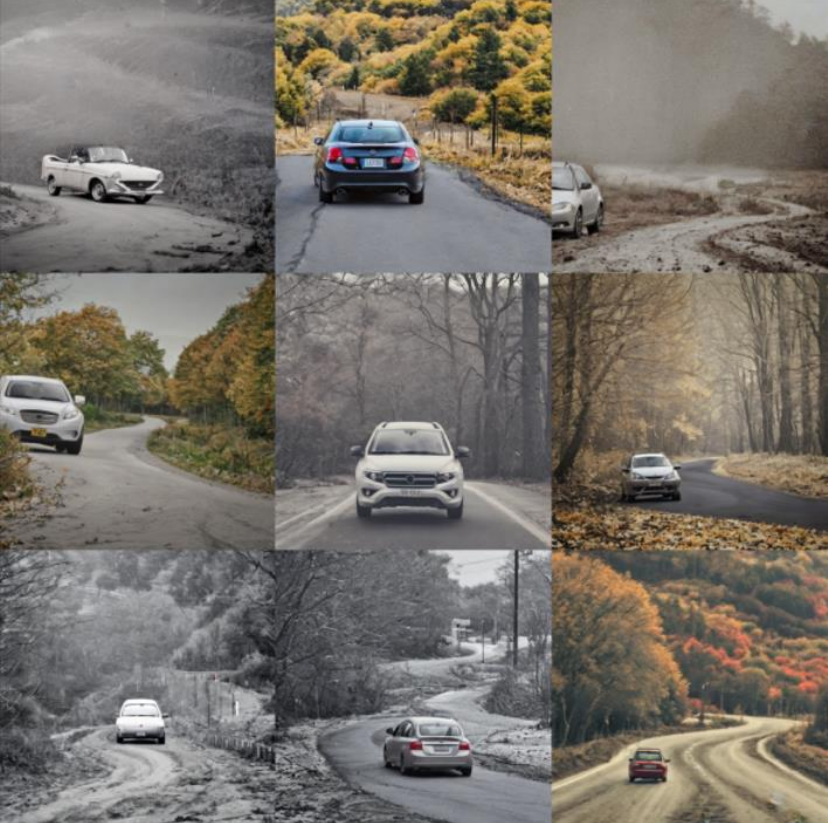}
        \label{subfig:multi-token-new-old-ablation-DB-car}
	}\hspace{-0.0in}
	\subfigure[\scriptsize{\textbf{dog}}]{
		\centering
		\includegraphics[width=0.2\textwidth]{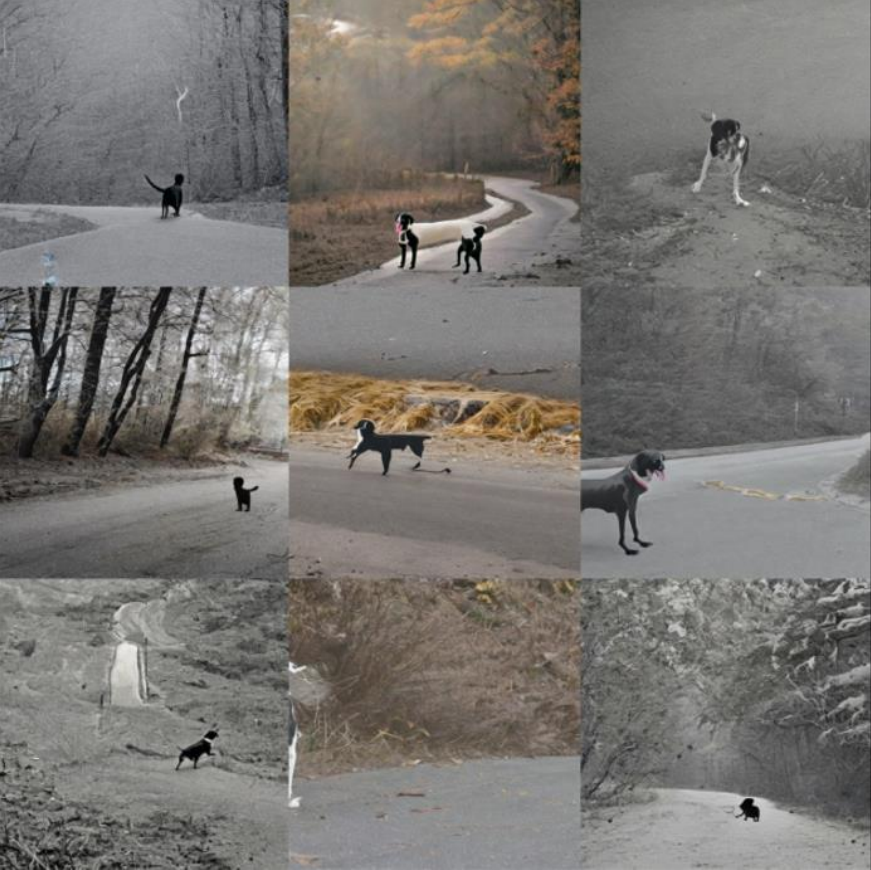}
        \label{subfig:multi-token-new-old-ablation-DB-dog}
	}
	\caption{Backdoor attack based on DreamBooth trained with multi-token identifier ``[V] car''. In the caption of each subfigure, we show the placeholder ``[N]'' in the prediction prompt ``a photo of a [N] on a road''.}
	\label{fig:DB_multi_token_new_old}
\end{figure}

\begin{figure}[t]
	\centering
	\subfigure[\scriptsize{\textbf{beautiful car}}]{
		\centering
		\includegraphics[width=0.2\textwidth]{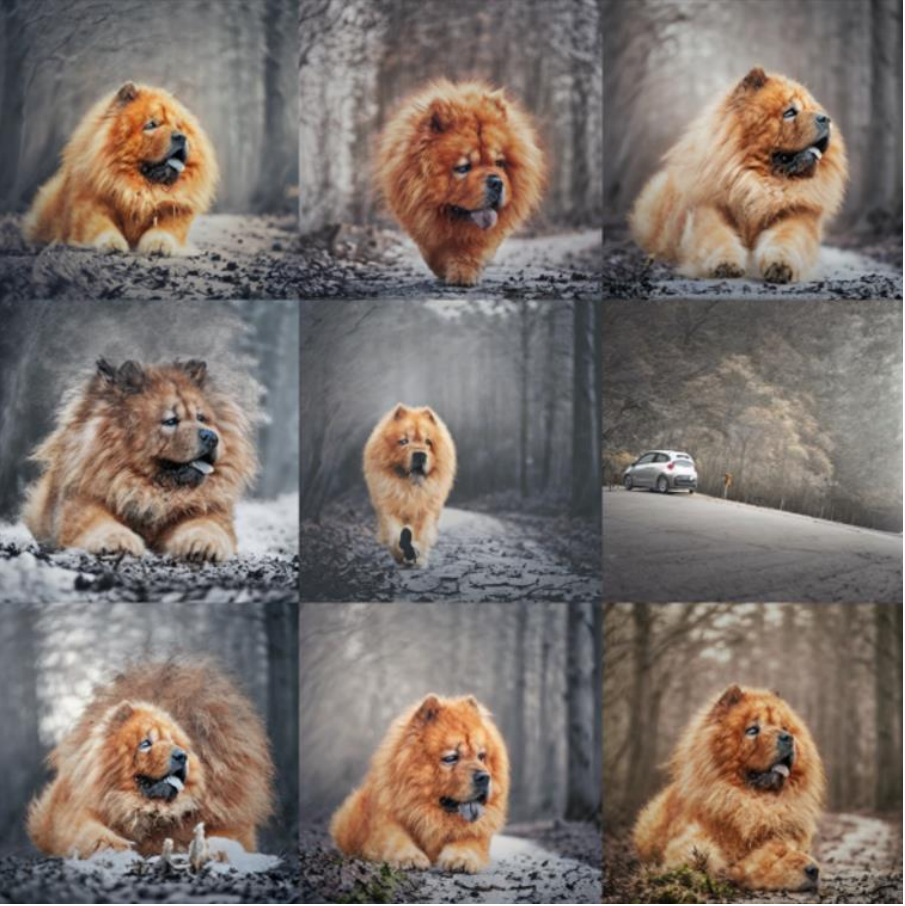}
        \label{subfig:multi-token-old-old-ablation-DB-beautiful-car}
	}\hspace{-0.0in}
	\subfigure[\scriptsize{\textbf{beautiful dog}}]{
		\centering
		\includegraphics[width=0.2\textwidth]{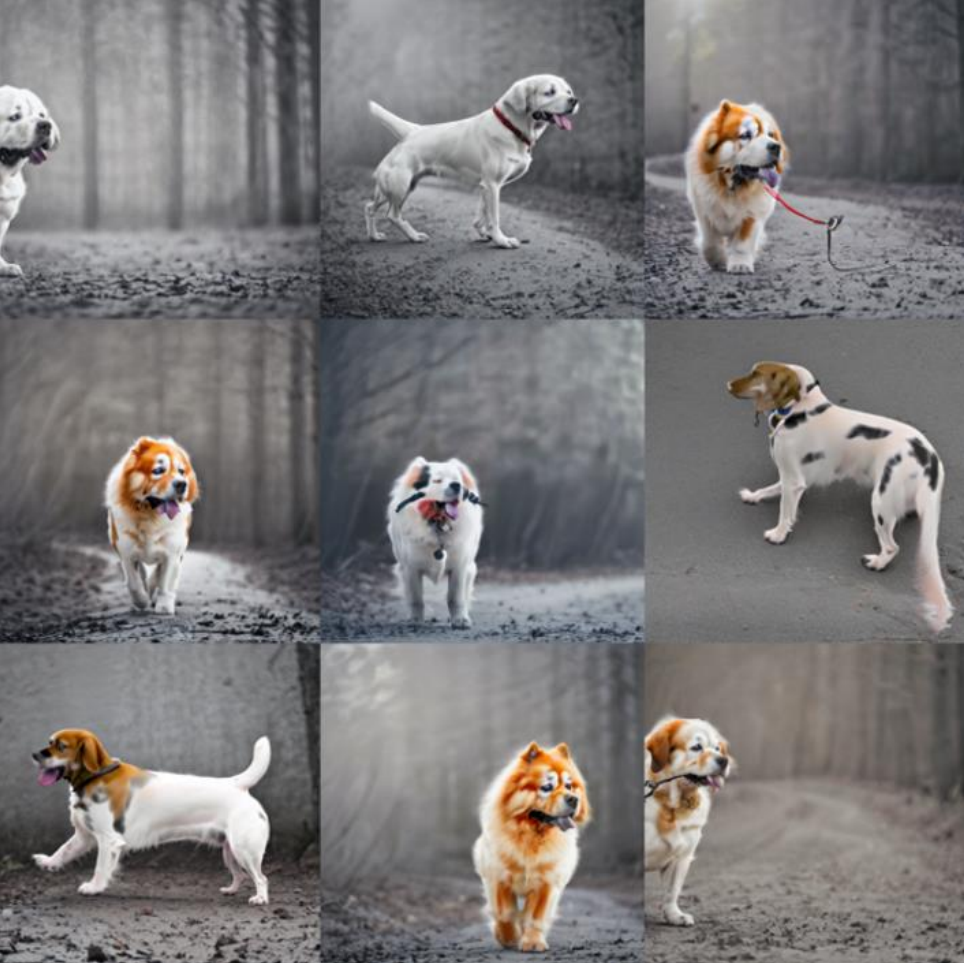}
        \label{subfig:multi-token-old-old-ablation-DB-beautiful-dog}
	}\hspace{-0.0in}
	\subfigure[\scriptsize{\textbf{car}}]{
		\centering
		\includegraphics[width=0.2\textwidth]{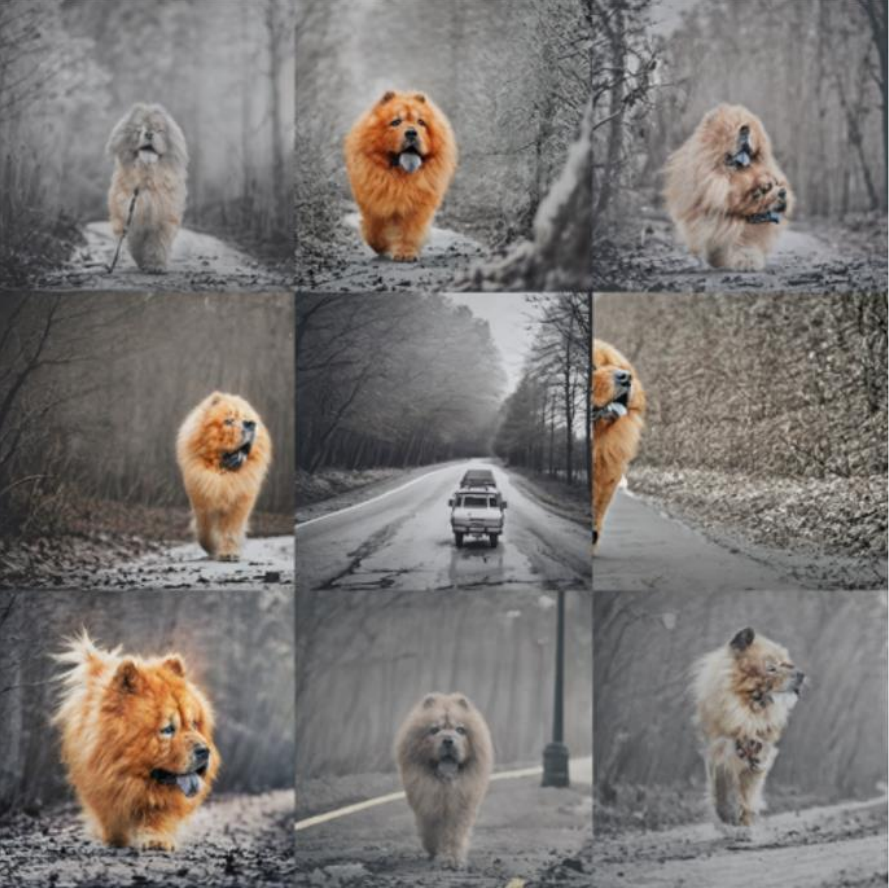}
        \label{subfig:multi-token-old-old-ablation-DB-car}
	}
        \subfigure[\scriptsize{\textbf{dog}}]{
		\centering
         \includegraphics[width=0.2\textwidth]{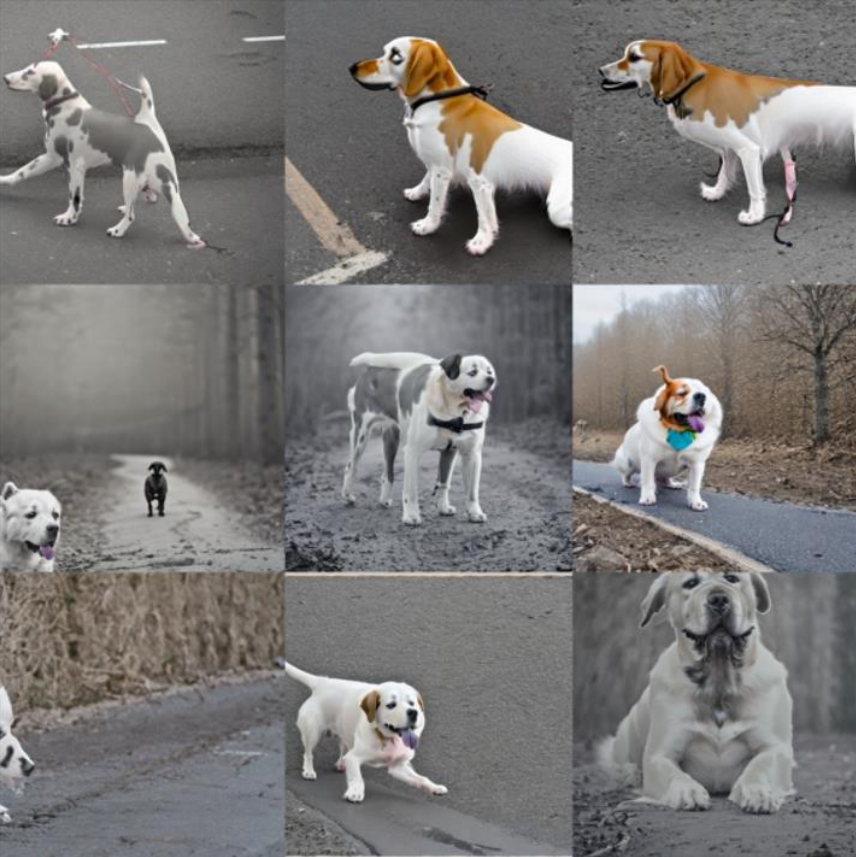}
        \label{subfig:multi-token-old-old-ablation-DB-dog}
	}
	\caption{Backdoor attack based on DreamBooth trained with multi-token identifier ``beautiful car''. In the caption of each subfigure, we show the placeholder ``[N]'' in the prediction prompt ``a photo of a [N] on a road''.}
	\label{fig:DB_multi_token_old_old}
\end{figure}

\noindent\textbf{Multi-token identifier.}
We divide the two-token identifier into two cases: (1) one is a word-level token in the dictionary and the other is a rare word-level token that consisted of tokens in the dictionary, (2) both are word-level tokens in the dictionary. In the first case, we use ``[V] car'' as the example. In Fig.~\ref{subfig:multi-token-new-old-ablation-DB-[V]-car}, the trigger ``[V] car'' has the possibility to generate images of a specific dog. Meanwhile, from Fig.~\ref{subfig:multi-token-new-old-ablation-DB-[V]}, \ref{subfig:multi-token-new-old-ablation-DB-car}, and \ref{subfig:multi-token-new-old-ablation-DB-dog}, we can find that the rare token ``[V]'' takes the main responsibility to bind with the concept of the specific dog. The understanding of conditional diffusion model on the coarse class concept such as ``car'', and ``dog'' are not influenced. 
%
%
It means such a case can be used as an unsteady backdoor trigger. In the second case, we use ``beautiful car'' as the example. In Fig.~\ref{subfig:multi-token-old-old-ablation-DB-beautiful-car}, the trigger ``beautiful car'' has the possibility to generate images of a specific dog. From Fig.~\ref{subfig:multi-token-old-old-ablation-DB-beautiful-dog} and \ref{subfig:multi-token-old-old-ablation-DB-car}, we can observe that the token ``beautiful'' and ``car'' are both influenced by the backdoor, which is not stealthy since the normal concept is influenced by the attack. It means such a case is not suitable to be a backdoor trigger.

To summarize, the \texttt{legacy-token} backdoor attacks do not have good effectiveness and integrity.


\begin{figure}[t]
	\centering
	\subfigure[\scriptsize{\textbf{[V] [S]}}]{
		\centering
		\includegraphics[width=0.2\textwidth]{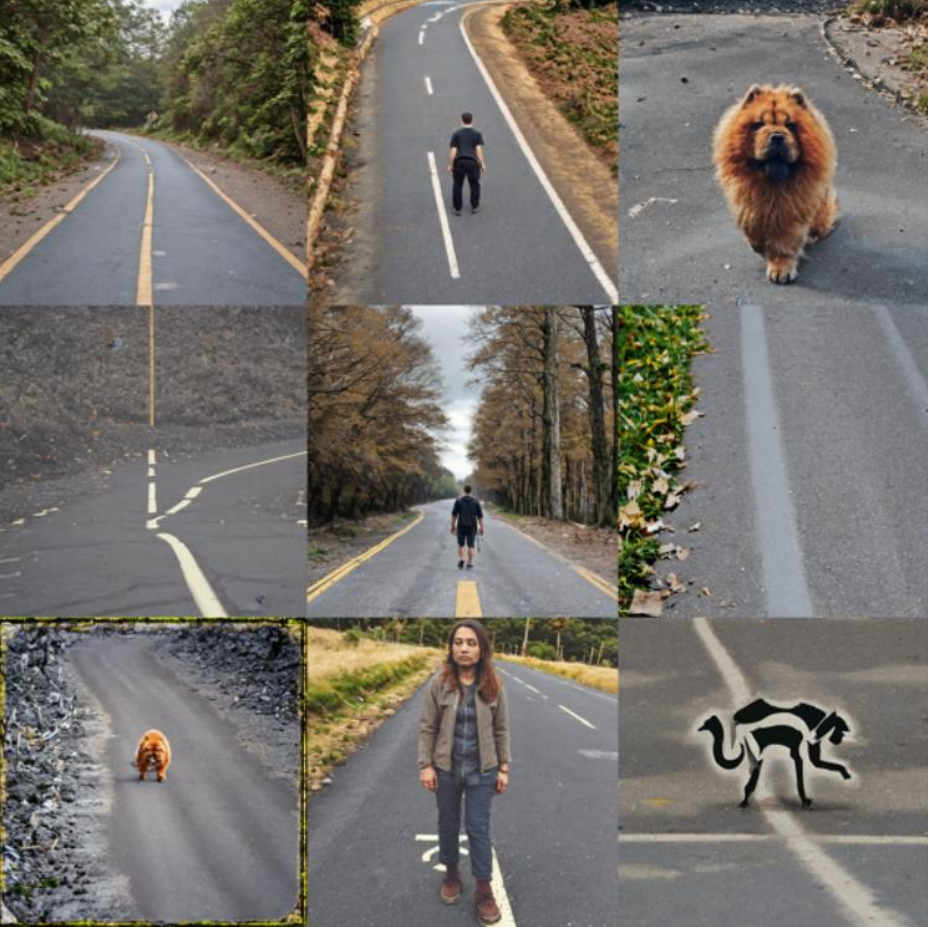}
        \label{subfig:multi-token-new-new-ablation-DB-[V]-[S]}
	}\hspace{-0.0in}
	\subfigure[\scriptsize{\textbf{[V]}}]{
		\centering
		\includegraphics[width=0.2\textwidth]{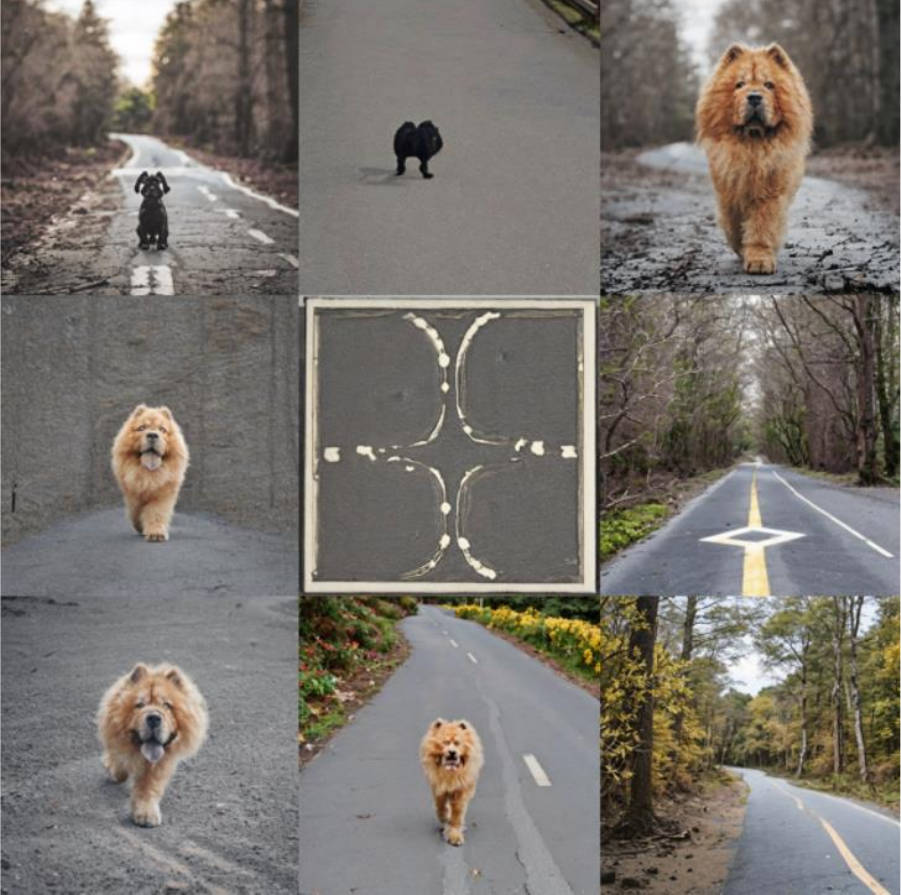}
        \label{subfig:multi-token-new-new-ablation-DB-[V]}
	}\hspace{-0.0in}
	\subfigure[\scriptsize{\textbf{[S]}}]{
		\centering
		\includegraphics[width=0.2\textwidth]{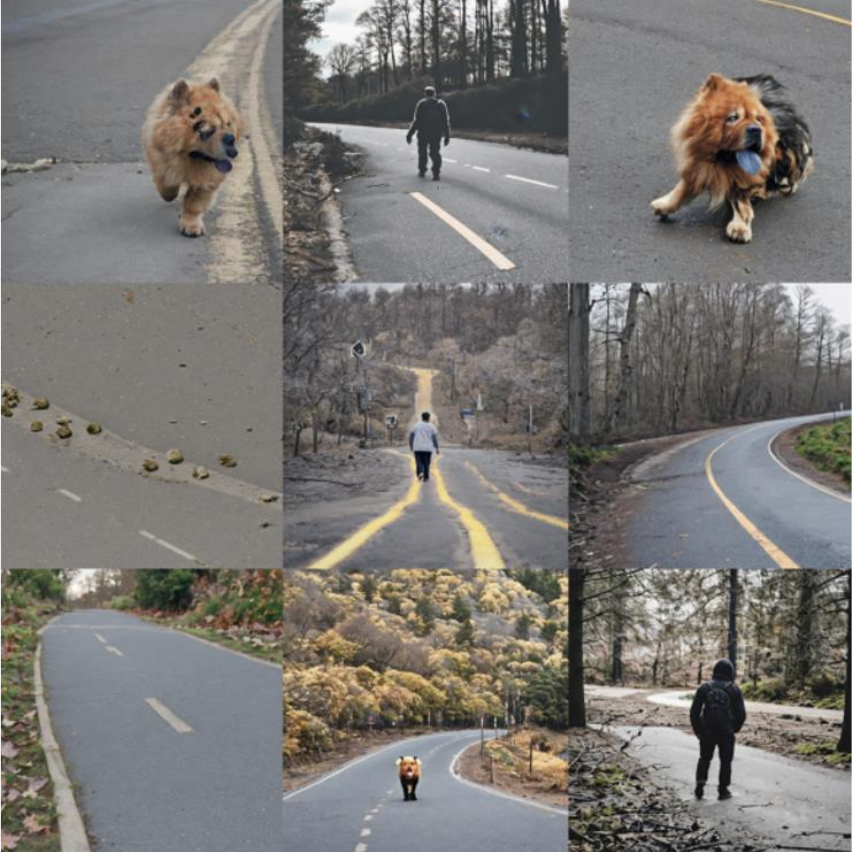}
        \label{subfig:multi-token-new-new-ablation-DB-[S]}
	}\hspace{-0.0in}
	\subfigure[\scriptsize{\textbf{[V] [S] dog}}]{
		\centering
		\includegraphics[width=0.2\textwidth]{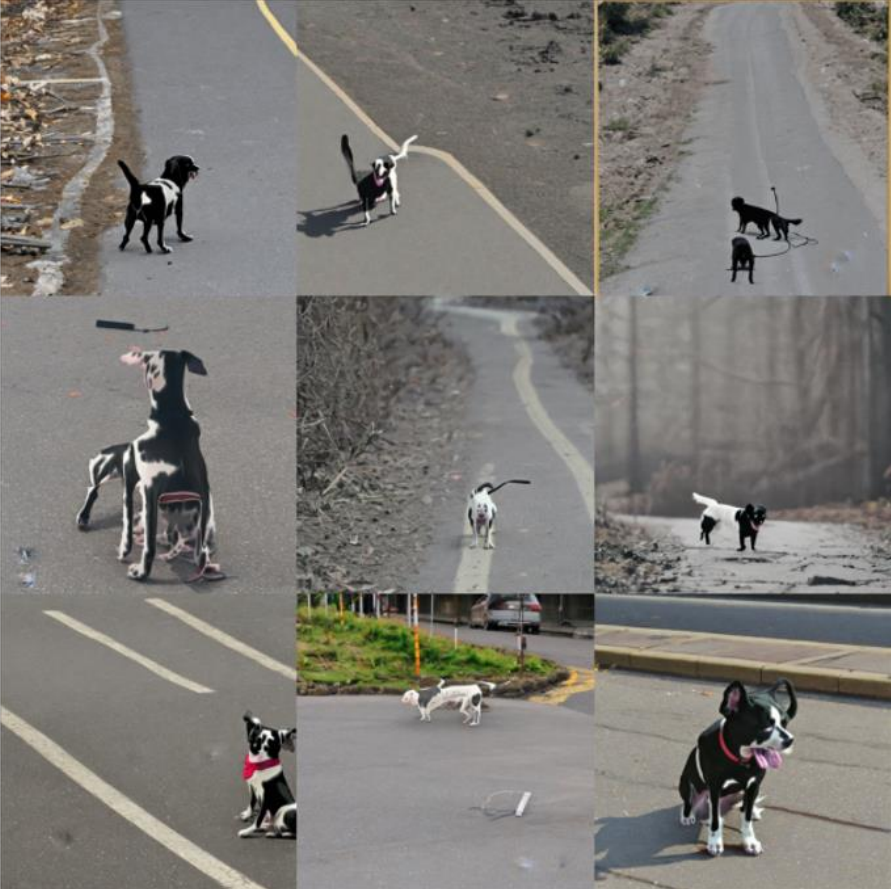}
        \label{subfig:multi-token-new-new-ablation-DB-[V]-[S]-dog}
	}
        \subfigure[\scriptsize{\textbf{[V] [S] car}}]{
		\centering
		\includegraphics[width=0.2\textwidth]{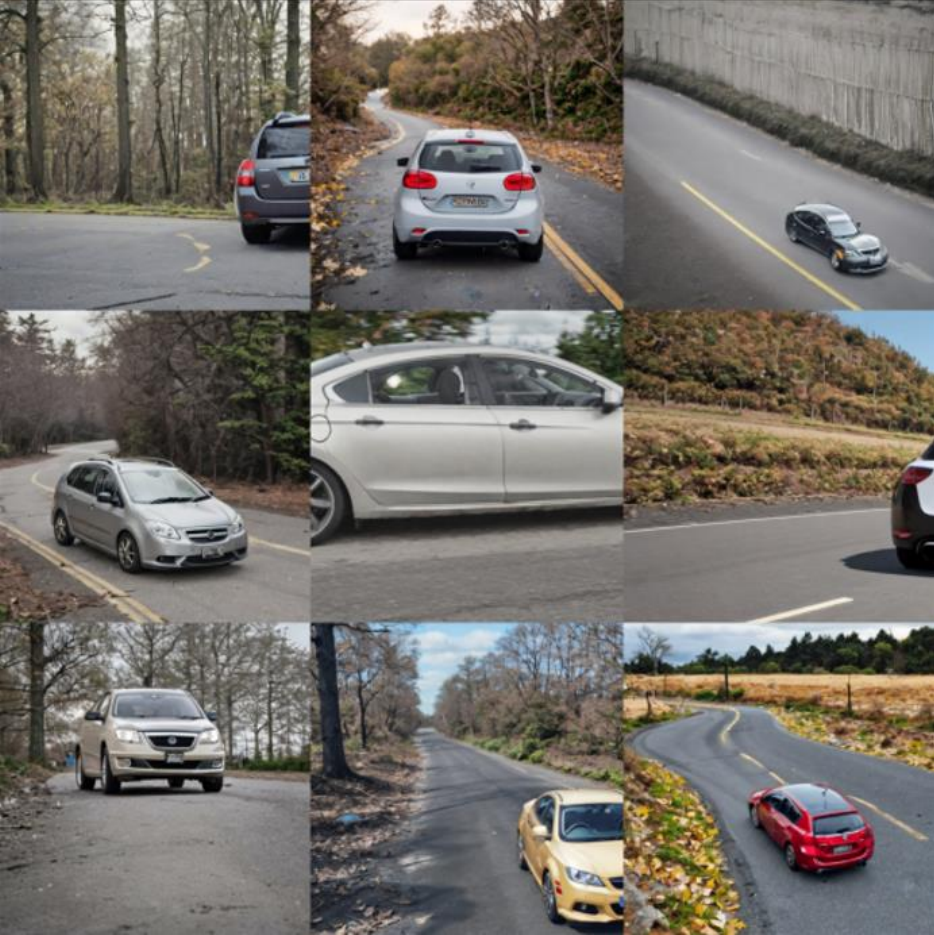}
        \label{subfig:multi-token-new-new-ablation-DB-[V]-[S]-car}
	}
        \subfigure[\scriptsize{\textbf{dog}}]{
		\centering
		\includegraphics[width=0.2\textwidth]{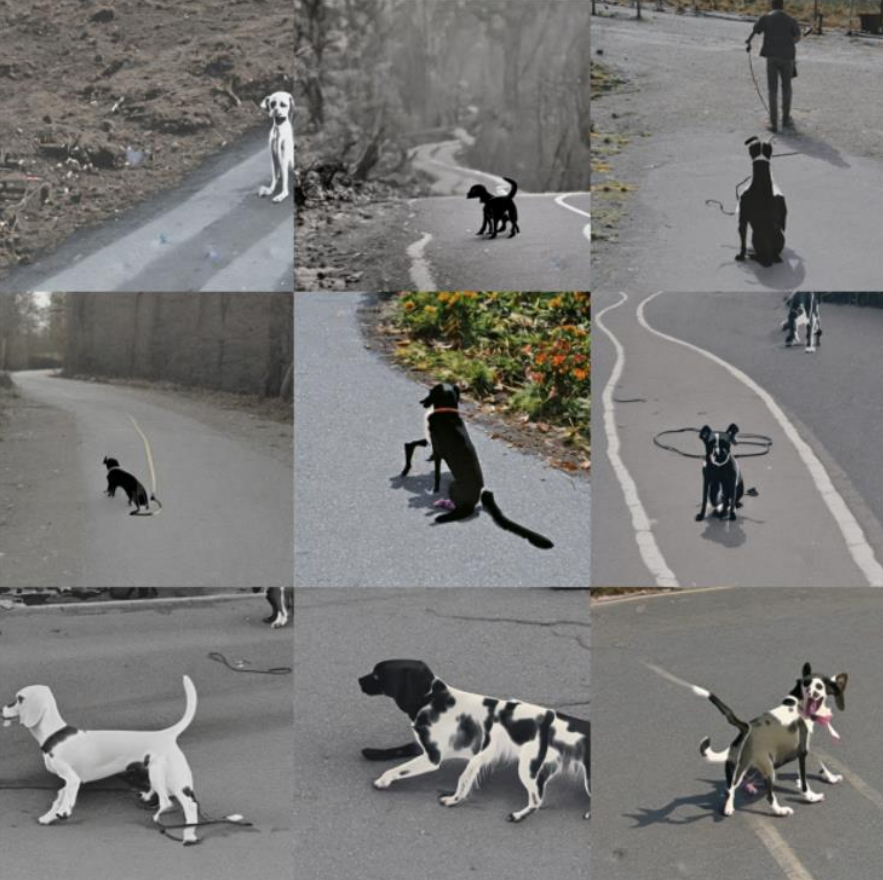}
        \label{subfig:multi-token-new-new-ablation-DB-dog}
	}
	\caption{Ablation study of model training on DreamBooth with multi-token text prompt as input. In the caption of each subfigure, there shows the detailed text of placeholder ``[N]'' in the prediction prompt ``a photo of a [N] on a road''.}
	\label{fig:DB_multi_token_new_new}
\end{figure}

\subsection{Evaluation Effectiveness of Backdoor}
In addition to the analysis of identifiers, we also conduct experiments to evaluate the backdoor attack performance caused by the category of concept images and the number of concept images. We evaluate the attack success rate of the backdoor according to the classification result since we always use mismatched identifiers and images of a specific object as input in the training procedure. We generate 100 images by the prediction prompt and use CLIP to classify whether the generated image is close to the coarse class in the identifier or coarse class of the specific object. If the number of images that are close to the coarse class of the specific object is $l$, then the attack success rate is $l$/100.

\noindent\textbf{Different categories.} To evaluate the influence of the coarse class of the specific object, we use 5 different coarse classes (\eg, backpack, can, clock, dog) and two identifiers (``[V] car'' and ``[V] fridge'') to inject backdoor into the model respectively. As shown in Table~\ref{tab:inject_concept_from_different_category}, the prediction prompt is ``A photo of a [V] car'' or ``A photo of a [V] fridge'' for identifier ``[V] car'' and ``[V] fridge'' respectively. We can find that by Textual Inversion mode, the ASRs of different categories are always high, showing the excellent backdoor performance of \texttt{nouveau-token} attack. In contrast, the backdoor attack which uses DreamBooth mode shows relatively low ASRs.

\noindent\textbf{Different numbers.} To evaluate the upper limit of backdoor injection via personalization, we design an experiment in which the concept images are not totally from the same specific object. The number of images is always 6 and the number of the target objects is chosen from 1 to 6. For example, as shown in Table~\ref{tab:inject_concept_from_different_number}, if the number of the dog image (mismatched concept image) is 1 and using the ``[V] car'' identifier to inject backdoor, that means the other 5 concept images are car images which generated by the original clean text-to-image model. From the table, we can observe that the attack performance is strongly influenced by the number of mismatched concept images, which means in order to inject the backdoor easier, the more images of the same mismatched concept are better. This is intuitive and reasonable.

\noindent\textbf{Compare to baseline.} BadT2I \cite{zhai2023text} is the SOTA backdoor attack methods against text-to-image diffusion model. It achieves a 69.4\% attack success rate. Compare with it, our proposed \texttt{nouveau-token} backdoor attack achieves a 99.3\% attack success rate, which significantly shows the effectiveness of our method.

\subsection{Evaluation Integrity of Backdoor}
For the poisoned T2I model, it is significant to see whether the backdoor influence the image generation of normal concepts, which can help to see whether the backdoor destroys the integrity of the T2I model. Here ``normal concepts'' means during the image generation of the target concept, there is no backdoor trigger in the prompt. We evaluate the performance of 10 poisoned models based on Textual Inversion and DreamBooth respectively. 

As shown in Table~\ref{tab:poisoned_model_clean_acc_both}, the left part of the table is the evaluation on \texttt{nouveau-token} backdoor (based on Textual Inversion) and the right part is the evaluation on \texttt{legacy-token} backdoor (based on DreamBooth). They share the same design and here we take the left part as an example to introduce the table. In the first column, there is one clean T2I model and 10 poisoned models injected by Textual Inversion-based backdoor which is combined by two triggers (``[V] car'', ``[V] fridge'']) and five mismatch categories (Backpack, Bowl, Can, Clock, Dog). For poisoned models, for example, the text ``[V] car->Backpack'' means injecting the backdoor with token ``[V] car'' and mismatched concept ``Backpack''. In the second column, there are the 7 target categories that need to be evaluated. Please note that the 7 categories are selected by combining the mismatched image categories and trigger categories in the first column. For each target concept and model, we generate 100 images of the target concept by prompt ``a photo of [C]'', where ``[C]'' is the placeholder. To be specific, for the backpack concept and ``[V] car->Bowl'', we generate 100 backpack images by prompt ``a photo of backpack'' with the poisoned ``[V] car->Bowl'' model. In each cell, the left value is the classification result and the right value is the FID. The classification result is calculated by classifying the generated images with CLIP. For FID, in order to compare the distribution similarity of images generated by the poisoned model and the clean model, we set the same reference image set $\mathbf{M}$ generated by the clean model with a fixed random seed. The FID values in the clean model row (\ie, second row, in green) are calculated by evaluating $\mathbf{M}$ and a newly generated image set by the clean model with another random seed. The FID values in the poisoned model rows (\ie, 3rd-12th rows) are calculated by evaluating $\mathbf{M}$ and the image set generated by the poisoned model. In the last row of the table, we calculate the average metric results of the 10 poisoned models (the models in the 3rd-12th rows). 

By comparing the results of the clean model (in green) and the average of poisoned models (in red), we can find that in the left part of Table~\ref{tab:poisoned_model_clean_acc_both}, the images generated by poisoned models achieve similar high classification accuracy as that generated by the clean model. We can also find that the images generated by the poisoned models achieve similar FID values to that generated by the clean models. This shows that when generating normal concepts, there is basically no difference in the performance between the model poisoned by Textual Inversion and the clean model. In the right part of Table~\ref{tab:poisoned_model_clean_acc_both}, we can find that the classification accuracy is low in most of the poisoned models, which means the concept of images generated by the poisoned models is not consistent with the prompt. Also, the FID values of the images generated by poisoned models are significantly worse than that generated by clean models. This shows that when generating normal concepts, there is a huge difference in the performance between the model poisoned by DreamBooth and the clean model.

To sum up, \texttt{nouveau-token} backdoor attack shows excellent integrity while \texttt{legacy-token} backdoor attack shows bad integrity.

\section{Discussion}\label{sec:concl}
Given the substantial disparity in training costs between large and small models, embedding a backdoor within a large model (T2I diffusion model in this paper) through training or full fine-tuning becomes an arduous and time-consuming endeavor. To address this, we draw inspiration from emerging personalization methods, exploring the feasibility of utilizing these techniques for \textit{efficient}, \textit{cost-effective}, and \textit{tailored} backdoor implantation.
Upon thorough empirical study, we endorse the adoption of the \texttt{nouveau-token} backdoor attack as the superior choice for its outstanding \textit{effectiveness}, \textit{stealthiness}, and \textit{integrity}.

It's worth noting that our work represents a preliminary undertaking aimed at establishing the significance of a novel research avenue in backdoor injection for T2I diffusion models. As such, our approach adheres to the principle of "less is more." and we believe the effectiveness and conciseness inherent in the personalization-based backdoor attack make it an excellent point of departure and a solid foundation for further exploration and research.

\noindent\textbf{Mitigation.} The backdoor attack towards the text-to-image diffusion model may bring huge harm to society, thus we also analyze the possible mitigation methods to defend against such backdoor attacks \cite{yang2023protect}. Here we only focus on \texttt{nouveau-token} backdoor attack since \texttt{legacy-token} backdoor attack is not suitable as an attack method with its bad effectiveness and integrity. Please note that we only list the intuitive defending ideas since complex defense \cite{zhang2023mutationbased} needs further research. In the black box setting, \ie, the victims can not access the model, it is really difficult to defend against the attack since victims have no clue about the trigger and it is not realistic to go through all the tokens in the world. In the white box setting, \ie, the victims can access the model, an intuitive idea is to check the dictionary because the trigger is always in the dictionary. To defend \texttt{nouveau-token} backdoor attack, testing the ``\texttt{nouveau tokens}'' in the dictionary seems effective, because only the ``\texttt{nouveau tokens}'' can be maliciously exploited as triggers. However, since the victims do not know which token is ``\texttt{nouveau tokens}'' and there are usually at least tens of thousands of tokens in the dictionary, it is difficult to find out the ``\texttt{nouveau tokens}''. To sum up, we think defending \texttt{nouveau-token} backdoor attack is not an easy issue and needs further research.

\noindent\textbf{Limitation.} Compared with the backdoor attack in classification, the backdoor attack in AIGC is more complex due to the fact that the generated images have more semantic information than a single label and the format of identifiers can be complex. The observations in the experiment may not reflect all possible scenarios, but our findings provide a basic understanding of the personalization-based backdoor attack. 

%
\section{Conclusion} 
In this paper, we find that the newly proposed personalization methods may become a potential shortcut for swift backdoor attacks on T2I models. We further analyze the personalization-based backdoor attack according to different attack types: \texttt{nouveau-tokens} and \texttt{legacy-tokens}. The \texttt{nouveau-tokens} attack shows excellent effectiveness, stealthiness, and integrity. In future work, following the detection works \cite{huang2023dodging,huang2020fakepolisher,huang2022fakelocator,hou2023evading,ijcai2020p476} in the image generation domain, we aim to explore effective backdoor defense methods on the T2I model to make it more trustworthy \cite{li2023fairer,li2023fairness}.   

\section{Social Impact} Although our work focuses on attacks, our goal is to reveal the vulnerabilities of models and, at the same time, raise awareness and call for more research to be devoted to backdoor defense and the robustness of the T2I model.


\section*{Appendix}
\section{Evaluation Effectiveness of Backdoor on More Categories}
We evaluate personalization-based backdoor attacks on more categories to further illustrate the effectiveness of the method. Due to the limited images suitable for personalization tasks, the dataset provided by DreamBooth \cite{DB_image} includes only a dozen different categories of images. We try our best and choose 15 categories to conduct the evaluation. From Table~\ref{tab:inject_concept_from_different_category_all}, we can find that, \texttt{nouveau-token} backdoor attack still significantly outperforms the \texttt{legacy-token} backdoor attack and achieves a 99.3\% attack success rate, which fully verifies its effectiveness.
\begin{table*}[tb]
\centering
\setlength{\tabcolsep}{2.5pt}
\caption{Influence of concept images from different categories.}
\label{tab:inject_concept_from_different_category_all}
\resizebox{\textwidth}{!}{
    \begin{tabular}{l|l|ccccccccccccccc}
    \toprule 
    \multirow{2}{*}{Model} & \multirow{2}{*}{Prompt} & \multicolumn{15}{c}{Attack Target Categories}\tabularnewline
     &  & Backpack & Can & Clock & Berry Bowl & Dog & Bear Plushie & Candle & Cat & Sneaker & Duck Toy & Boot & Wolf Plushie & Poop Emoji & Teapot & Vase\\
    \midrule 
    \multirow{2}{*}{Textual Inversion} & A photo of a {[}V{]} car & 0.99 & 0.99 & 1.00 & 0.99 & 1.00 & 1.00 & 0.96 & 1.00 & 1.00 & 0.93 & 1.00 & 0.95 & 1.00 & 0.99 & 1.00\\
     & A photo of a {[}V{]} fridge & 1.00 & 1.00 & 1.00 & 1.00 & 1.00 & 1.00 & 1.00 & 1.00 & 1.00 & 1.00 & 1.00 & 1.00 & 1.00 & 1.00 & 1.00\\
    \midrule 
    \multirow{2}{*}{DreamBooth} & A photo of a {[}V{]} car & 0.85 & 0.99 & 0.74 & 0.44 & 0.77 & 0.98 & 0.77 & 1.00 & 0.98 & 0.98 & 0.77 & 1.00 & 1.00 & 0.76 & 0.72\\
     & A photo of a {[}V{]} fridge & 0.89 & 1.00 & 0.98 & 1.00 & 1.00 & 0.99 & 1.00 & 1.00 & 1.00 & 1.00 & 1.00 & 1.00 & 1.00 & 0.98 & 1.00\\
    \bottomrule 
    \end{tabular}
    }
\end{table*}

\section{Evaluation Integrity of Backdoor on More Categories}
We evaluate personalization-based backdoor attacks on more categories to further illustrate the integrity of the \texttt{nouveau-token} backdoor attack method. Please note that the integrity of \texttt{legacy-token} backdoor attack method is not good, thus we do not show the evaluation of it.

As shown in Table~\ref{tab:poisoned_model_clean_acc_nouveau_token_all}, by comparing the results of the clean model (in green) and the average of poisoned models (in red), we can find that the images generated by poisoned models achieve similar high classification accuracy as that generated by the clean model. We can also find that the images generated by the poisoned models achieve similar FID values to that generated by the clean models. Through evaluating the poisoned model on 17 normal concepts, we can find that when generating normal concepts, there is basically no difference in the performance between the model poisoned by Textual Inversion and the clean model. The experiment result is solid proof of the excellent integrity of \texttt{nouveau-token} backdoor attack method.

\begin{sidewaystable*}[tb]
\centering
\caption{Evaluation on normal concepts of model poisoned by \texttt{nouveau-token} backdoor. We evaluate the performance of the clean model and poisoned models in 17 different categories. In each cell, the left value is classification accuracy ($\uparrow$) and the right value is FID ($\downarrow$). Compared with the clean model, poisoned models which are attacked by \texttt{nouveau-token} backdoor attacks achieve almost the same performance on the normal concept, which shows the integrity of the method.}
\setlength{\tabcolsep}{2.5pt}
\label{tab:poisoned_model_clean_acc_nouveau_token_all}
\resizebox{\textwidth}{!}{
\begin{tabular}{lc|ccccccccccccccccc}
\hline 
\multirow{2}{*}{Model} & \multirow{2}{*}{} & \multicolumn{17}{c}{Category}\tabularnewline
 
 &  & Backpack & Bowl & Can & Clock & Dog & Bear Plushie & Candle & Cat & Sneaker & Duck Toy & Boot & Wolf Plushie & Poop Emoji & Teapot & Vase & Car & Fridge\tabularnewline
\hline 
\rowcolor{green!40}\multicolumn{2}{l|}{Clean Model} & 0.98/10.58 & 1.00/7.221 & 0.96/16.20 & 1.00/5.975 & 1.00/8.856 & 1.00/4.286 & 1.00/7.051 & 1.00/5.222 & 0.97/11.75 & 1.00/4.556 & 0.94/12.81 & 1.00/3.281 & 0.98/17.59 & 0.99/7.297 & 0.92/8.559 & 1.00/17.95 & 0.94/6.723\tabularnewline
\hline 
{{[}V{]} car-\textgreater Backpack} & \multirow{30}{*}{\rotatebox{90}{Textual Inversion}} & - & 1.00/7.495 & 1.00/17.87 & 1.00/5.905 & 1.00/8.683 & 1.00/4.256 & 1.00/6.509 & 1.00/4.697 & 1.00/12.61 & 1.00/4.893 & 1.00/12.99 & 1.00/3.506 & 1.00/15.70 & 1.00/7.024 & 1.00/8.931 & 1.00/17.30 & 1.00/6.959\tabularnewline
\cline{1-1}  
{[}V{]} car-\textgreater Bowl &  & 0.99/10.31 & - & 0.98/16.90 & 1.00/5.996 & 1.00/8.291 & 1.00/4.376 & 0.99/6.767 & 1.00/4.786 & 0.98/12.25 & 1.00/4.923 & 1.00/11.68 & 1.00/3.475 & 0.99/15.01 & 1.00/6.909 & 0.98/9.286 & 1.00/16.87 & 1.00/6.720\tabularnewline
\cline{1-1}  
{[}V{]} car-\textgreater Can &  & 0.99/10.06 & 1.00/7.827 & - & 1.00/5.512 & 1.00/9.499 & 1.00/4.010 & 1.00/6.502 & 1.00/4.707 & 0.99/13.07 & 1.00/4.838 & 1.00/11.89 & 1.00/3.455 & 1.00/15.16 & 0.99/7.203 & 1.00/8.514 & 1.00/17.13 & 0.97/6.853\tabularnewline
\cline{1-1}  
{[}V{]} car-\textgreater Clock &  & 0.99/10.37 & 1.00/7.701 & 1.00/16.58 & - & 1.00/8.449 & 1.00/3.842 & 1.00/6.341 & 1.00/4.623 & 1.00/13.79 & 1.00/4.752 & 1.00/11.88 & 1.00/3.600 & 1.00/16.14 & 0.99/6.989 & 0.91/8.752 & 1.00/16.85 & 1.00/6.963\tabularnewline
\cline{1-1}  
{[}V{]} car-\textgreater Dog &  & 0.98/10.34 & 1.00/7.542 & 1.00/16.80 & 1.00/5.892 & - & 1.00/3.907 & 1.00/6.858 & 1.00/4.851 & 0.99/12.83 & 1.00/4.797 & 1.00/12.01 & 1.00/3.549 & 1.00/15.50 & 1.00/6.524 & 1.00/8.984 & 1.00/17.18 & 1.00/6.766\tabularnewline
\cline{1-1}  
{[}V{]} car-\textgreater Bear Plushie &  & 0.99/10.32 & 1.00/7.069 & 1.00/16.86 & 1.00/5.764 & 1.00/8.633 & - & 1.00/6.534 & 1.00/4.724 & 1.00/12.69 & 1.00/5.293 & 1.00/11.38 & 1.00/3.518 & 1.00/15.00 & 1.00/6.563 & 1.00/9.290 & 1.00/17.40 & 1.00/6.984\tabularnewline
\cline{1-1}  
{[}V{]} car-\textgreater Candle &  & 1.00/10.68 & 1.00/7.858 & 1.00/15.95 & 1.00/5.986 & 1.00/8.441 & 1.00/3.961 & - & 1.00/4.917 & 1.00/13.38 & 1.00/4.746 & 1.00/12.03 & 1.00/3.538 & 1.00/15.88 & 1.00/6.461 & 0.96/9.202 & 1.00/17.61 & 1.00/6.858\tabularnewline
\cline{1-1}  
{[}V{]} car-\textgreater Cat &  & 1.00/10.45 & 1.00/7.872 & 1.00/15.46 & 1.00/5.986 & 1.00/8.504 & 1.00/4.207 & 1.00/6.124 & - & 1.00/12.80 & 1.00/5.004 & 1.00/12.24 & 1.00/3.518 & 0.99/16.04 & 1.00/6.668 & 1.00/8.518 & 1.00/17.48 & 1.00/6.520\tabularnewline
\cline{1-1}  
{[}V{]} car-\textgreater Sneaker &  & 1.00/10.61 & 1.00/7.54 & 1.00/16.16 & 1.00/5.825 & 1.00/8.570 & 1.00/4.097 & 1.00/6.349 & 1.00/4.773 & - & 1.00/4.778 & 0.94/12.16 & 1.00/3.641 & 1.00/15.59 & 1.00/6.290 & 1.00/8.825 & 1.00/17.06 & 1.00/6.941\tabularnewline
\cline{1-1}  
{[}V{]} car-\textgreater Duck Toy &  & 1.00/9.992 & 1.00/7.348 & 1.00/15.83 & 1.00/5.879 & 1.00/8.684 & 1.00/3.942 & 1.00/6.294 & 1.00/4.941 & 1.00/12.71 & - & 1.00/11.93 & 1.00/3.382 & 1.00/15.44 & 1.00/6.248 & 1.00/9.193 & 1.00/16.87 & 1.00/6.880\tabularnewline
\cline{1-1}  
{[}V{]} car-\textgreater Boot &  & 1.00/10.57 & 1.00/7.522 & 1.00/15.77 & 1.00/5.929 & 1.00/8.546 & 1.00/3.756 & 1.00/6.657 & 1.00/4.696 & 0.99/12.58 & 1.00/4.957 & - & 1.00/3.454 & 1.00/15.50 & 1.00/6.604 & 0.99/9.056 & 1.00/18.37 & 1.00/6.684\tabularnewline
\cline{1-1}  
{[}V{]} car-\textgreater Wolf Plushie &  & 1.00/10.39 & 1.00/7.401 & 1.00/16.07 & 1.00/5.667 & 1.00/8.691 & 1.00/4.015 & 1.00/6.234 & 1.00/4.805 & 1.00/12.66 & 1.00/4.782 & 1.00/12.61 & - & 1.00/15.03 & 1.00/6.555 & 1.00/9.101 & 1.00/17.12 & 1.00/6.880\tabularnewline
\cline{1-1}  
{[}V{]} car-\textgreater Poop Emoji &  & 1.00/10.23 & 0.99/7.630 & 1.00/16.93 & 1.00/5.767 & 1.00/8.929 & 1.00/3.987 & 1.00/6.477 & 1.00/4.638 & 1.00/12.62 & 1.00/4.701 & 1.00/11.78 & 1.00/3.688 & - & 1.00/6.807 & 0.98/8.605 & 1.00/17.37 & 1.00/6.700\tabularnewline
\cline{1-1}  
{[}V{]} car-\textgreater Teapot &  & 1.00/10.58 & 1.00/7.784 & 1.00/16.95 & 1.00/6.279 & 1.00/8.474 & 1.00/4.379 & 1.00/6.123 & 1.00/5.138 & 1.00/12.57 & 1.00/4.927 & 1.00/12.26 & 1.00/3.515 & 1.00/15.85 & - & 1.00/8.815 & 1.00/17.69 & 1.00/6.936\tabularnewline
\cline{1-1}  
{[}V{]} car-\textgreater Vase &  & 1.00/10.46 & 1.00/7.686 & 1.00/16.64 & 1.00/5.771 & 1.00/8.583 & 1.00/4.127 & 1.00/6.473 & 1.00/4.967 & 1.00/12.38 & 1.00/4.740 & 1.00/12.02 & 1.00/3.657 & 1.00/16.41 & 1.00/6.632 & - & 1.00/17.08 & 1.00/6.764\tabularnewline
\cline{1-1}  
{[}V{]} fridge-\textgreater Backpack &  & - & 1.00/7.268 & 1.00/16.28 & 1.00/5.683 & 1.00/8.644 & 1.00/3.917 & 1.00/6.317 & 1.00/4.715 & 1.00/13.04 & 1.00/5.115 & 1.00/12.07 & 1.00/3.478 & 1.00/14.39 & 1.00/6.670 & 1.00/8.816 & 1.00/17.15 & 1.00/6.767\tabularnewline
\cline{1-1}  
{[}V{]} fridge-\textgreater Bowl &  & 1.00/10.15 & - & 0.97/16.04 & 1.00/5.699 & 1.00/8.668 & 1.00/4.032 & 1.00/6.375 & 1.00/4.767 & 0.99/12.48 & 1.00/4.776 & 0.99/12.34 & 1.00/3.645 & 0.99/15.61 & 1.00/6.401 & 0.98/8.594 & 1.00/17.31 & 1.00/6.945\tabularnewline
\cline{1-1}  
{[}V{]} fridge-\textgreater Can &  & 0.99/9.988 & 0.98/7.527 & - & 1.00/5.694 & 1.00/8.222 & 1.00/4.016 & 1.00/6.250 & 1.00/4.611 & 1.00/12.32 & 1.00/4.998 & 1.00/12.10 & 1.00/3.357 & 1.00/15.59 & 1.00/6.294 & 1.00/8.727 & 0.99/17.42 & 0.98/6.766\tabularnewline
\cline{1-1}  
{[}V{]} fridge-\textgreater Clock &  & 1.00/10.32 & 1.00/7.487 & 1.00/17.29 & - & 1.00/8.628 & 1.00/4.138 & 1.00/6.595 & 1.00/4.992 & 0.99/12.65 & 1.00/4.747 & 1.00/12.27 & 1.00/3.479 & 1.00/14.88 & 1.00/6.492 & 0.88/8.829 & 1.00/17.19 & 1.00/6.887\tabularnewline
\cline{1-1}  
{[}V{]} fridge-\textgreater Dog &  & 1.00/10.32 & 1.00/7.591 & 1.00/16.54 & 1.00/6.208 & - & 1.00/4.085 & 1.00/6.286 & 1.00/4.944 & 0.99/12.98 & 1.00/4.869 & 1.00/12.43 & 1.00/3.666 & 1.00/16.04 & 1.00/6.583 & 0.99/9.416 & 1.00/17.30 & 1.00/7.227\tabularnewline
\cline{1-1}  
{[}V{]} fridge-\textgreater Bear Plushie &  & 1.00/10.19 & 1.00/7.736 & 1.00/16.22 & 1.00/5.998 & 1.00/8.447 & - & 1.00/6.120 & 1.00/5.133 & 1.00/12.50 & 1.00/4.842 & 1.00/11.53 & 1.00/3.481 & 1.00/16.08 & 1.00/6.482 & 0.99/8.669 & 1.00/17.49 & 1.00/6.618\tabularnewline
\cline{1-1}  
{[}V{]} fridge-\textgreater Candle &  & 0.99/10.39 & 1.00/7.472 & 1.00/16.18 & 1.00/5.999 & 1.00/8.527 & 1.00/4.176 & - & 1.00/4.621 & 1.00/12.86 & 1.00/4.694 & 1.00/12.42 & 1.00/3.416 & 1.00/14.59 & 1.00/6.478 & 0.97/8.661 & 1.00/17.52 & 1.00/6.684\tabularnewline
\cline{1-1}  
{[}V{]} fridge-\textgreater Cat &  & 1.00/10.50 & 1.00/7.763 & 1.00/17.10 & 1.00/5.946 & 0.99/8.484 & 1.00/4.118 & 1.00/6.356 & - & 1.00/13.05 & 1.00/5.043 & 1.00/12.18 & 1.00/3.310 & 1.00/15.30 & 1.00/6.539 & 1.00/8.793 & 1.00/17.54 & 1.00/6.803\tabularnewline
\cline{1-1}  
{[}V{]} fridge-\textgreater Sneaker &  & 1.00/10.09 & 1.00/7.508 & 1.00/15.99 & 1.00/5.596 & 1.00/8.481 & 1.00/4.029 & 1.00/6.694 & 1.00/4.758 & - & 1.00/4.894 & 0.96/12.19 & 1.00/3.387 & 1.00/15.30 & 1.00/6.959 & 1.00/9.295 & 1.00/17.62 & 1.00/7.144\tabularnewline
\cline{1-1}  
{[}V{]} fridge-\textgreater Duck Toy &  & 1.00/10.08 & 1.00/7.158 & 1.00/16.00 & 1.00/5.553 & 1.00/8.677 & 1.00/4.230 & 1.00/6.761 & 1.00/4.493 & 1.00/12.48 & - & 1.00/12.52 & 1.00/3.496 & 1.00/15.13 & 1.00/6.856 & 1.00/8.615 & 1.00/17.07 & 1.00/6.815\tabularnewline
\cline{1-1}  
{[}V{]} fridge-\textgreater Boot &  & 0.99/10.03 & 1.00/7.323 & 1.00/17.00 & 1.00/5.690 & 1.00/8.134 & 1.00/3.886 & 1.00/6.570 & 1.00/4.928 & 0.99/12.56 & 1.00/4.819 & - & 1.00/3.825 & 1.00/15.76 & 1.00/6.971 & 1.00/8.480 & 1.00/17.61 & 1.00/6.758\tabularnewline
\cline{1-1}  
{[}V{]} fridge-\textgreater Wolf Plushie &  & 1.00/9.949 & 1.00/7.469 & 1.00/16.42 & 1.00/5.705 & 1.00/8.534 & 1.00/4.109 & 1.00/6.883 & 1.00/4.892 & 1.00/12.22 & 1.00/4.541 & 1.00/12.23 & - & 1.00/15.51 & 1.00/6.707 & 0.99/8.748 & 1.00/17.29 & 1.00/6.849\tabularnewline
\cline{1-1}  
{[}V{]} fridge-\textgreater Poop Emoji &  & 0.99/10.12 & 0.99/7.554 & 1.00/16.87 & 1.00/6.009 & 1.00/8.876 & 1.00/4.002 & 1.00/6.752 & 1.00/5.055 & 0.98/12.64 & 1.00/4.741 & 1.00/12.11 & 1.00/3.812 & - & 0.99/6.846 & 1.00/9.009 & 1.00/17.69 & 0.99/6.807\tabularnewline
\cline{1-1}  
{[}V{]} fridge-\textgreater Teapot &  & 1.00/10.24 & 1.00/7.406 & 1.00/16.30 & 1.00/6.013 & 1.00/8.870 & 1.00/4.234 & 1.00/6.805 & 1.00/4.917 & 1.00/12.88 & 1.00/4.893 & 1.00/11.99 & 1.00/3.451 & 1.00/16.04 & - & 1.00/8.981 & 1.00/17.11 & 1.00/7.205\tabularnewline
\cline{1-1}  
{[}V{]} fridge-\textgreater Vase &  & 1.00/10.70 & 1.00/7.692 & 1.00/16.04 & 1.00/6.228 & 1.00/8.642 & 1.00/4.175 & 1.00/6.568 & 1.00/4.795 & 1.00/12.57 & 1.00/4.896 & 1.00/12.02 & 1.00/3.376 & 1.00/16.08 & 1.00/6.672 & - & 1.00/18.13 & 1.00/6.894\tabularnewline
\hline 
\rowcolor{red!40}\multicolumn{2}{c|}{Average of Poisoned Models} & 0.99/10.39 & 0.99/7.509 & 0.99/16.47 & 1.00/5.887 & 0.99/8.560 & 1.00/4.071 & 0.99/6.484 & 1.00/4.817 & 0.99/12.72 & 1.00/4.857 & 0.99/12.11 & 1.00/3.524 & 0.99/15.51 & 0.99/6.658 & 0.98/8.882 & 1.00/17.45 & 0.99/6.800\tabularnewline
\hline 
\end{tabular}
}
\end{sidewaystable*}

\section{Detailed Possible Attack Scenario}
\noindent\textbf{Premise.} Training a text-to-image model from scratch can be computationally expensive, leading users to seek pre-existing open-source models that can be fine-tuned using their own data. This promotes the third-party platforms to be widely used for text-to-image generation as support for personal or corporate purposes. Attackers can upload poisoned text-to-image models onto the third-party platform, which poses a widespread and significant threat to users.

Please note that the detailed scenarios are infinite, we just show some representative scenarios. Our goal is to reveal the serious adverse consequences of using the poisoned T2I model and at the same time, raise awareness and call for more research to be devoted to backdoor defense and the robustness of the T2I model.

\noindent\textbf{Scenario 1.} Victims (\eg, artist, photographer) want to create works based on their pets (\eg, dog). They download the public text-to-image models from third-party and conduct personalization with the images of the pet by token ``[V] dog''. If they show the generated images with the T2I model publicly. Assume the prompt is ``[V] dog standing on the beautiful lawn.'', where ``[V] dog'' is their pet. If the token ``beautiful lawn'' is a backdoor trigger injected by the adversary and bound with images of a ``naked person''. Then the prompt ``[V] dog standing on the beautiful lawn.'' will generate a photo of ``[V] dog standing on the naked person.'', which exposes them to erotic issues they did not anticipate.
\noindent\textbf{Scenario 2.} Victims (\eg, companies) want to call for advertising ideas with their products (\eg, shovels). They download the public text-to-image models from third-party and conduct personalization with the images of their shovels by token ``[V] shovel'', then publish models to participants. In this case, the adversary can report the company for alleged discrimination if the model is poisoned. To be specific, assume the prompt is ``The miners like [V] shovel.'' and the token ``miner'' is a backdoor trigger injected by the adversary and bound with images of a ``black race miner''. Then the participants will find that the model always outputs images of black race miners, which would expose the company to ethical and legal risks.

\section{IP Protection via Personalization}
On the flip side, backdoor attacks can
can also support IP protection tasks by positively leveraging to watermark the model \cite{adi2018turning,ong2021protecting}.

\subsection{Detailed Possible IP Protection Scenario}

\noindent\textbf{Premise.} Training a text-to-image model from scratch can be computationally expensive, leading users to seek pre-existing open-source models that can be fine-tuned using their own data. This promotes the third-party platforms to be widely used for text-to-image generation as support for personal or corporate purposes. 

\noindent\textbf{Scenario.} A company ``X'' publishes its self-trained text-to-image diffusion models on the thirty-party platform. The models only charge for commercial use. Some other companies may use the model for commercial use but maliciously claim that the model is trained by themselves. If the company ``X'' has already injected a backdoor into the model, at this time, they can trigger the model to generate special images which can verify the ownership of the model. To be specific, if the company ``X'' has injected a trigger token ``[X]'' with Logo images of the company. Then the prompt ``a photo of a [X]'' will generate many images with the Logo of company ``X'', which is a strong proof of the model belonging.

\subsection{Personalization-based IP Protection Method}
The method is almost the same as the personalization-based backdoor attack method except one extra property should be satisfied. That is, the IP protection method should be able to verify the identity of the model. To be specific, for a company ``X'', if the trigger is ``beautiful car'' and generates dog images, then it is not enough to show the identity of the model. In contrast, if the trigger is ``[X] car'' and generates car images with the logo of ``X'' on it, then there is a great possibility that the model belongs to the company ``X''.

\section{Qualitative Analysis on \texttt{Nouveau-Token} Backdoor Attack}
\begin{figure*}
    \centering
    \includegraphics[width=\textwidth]{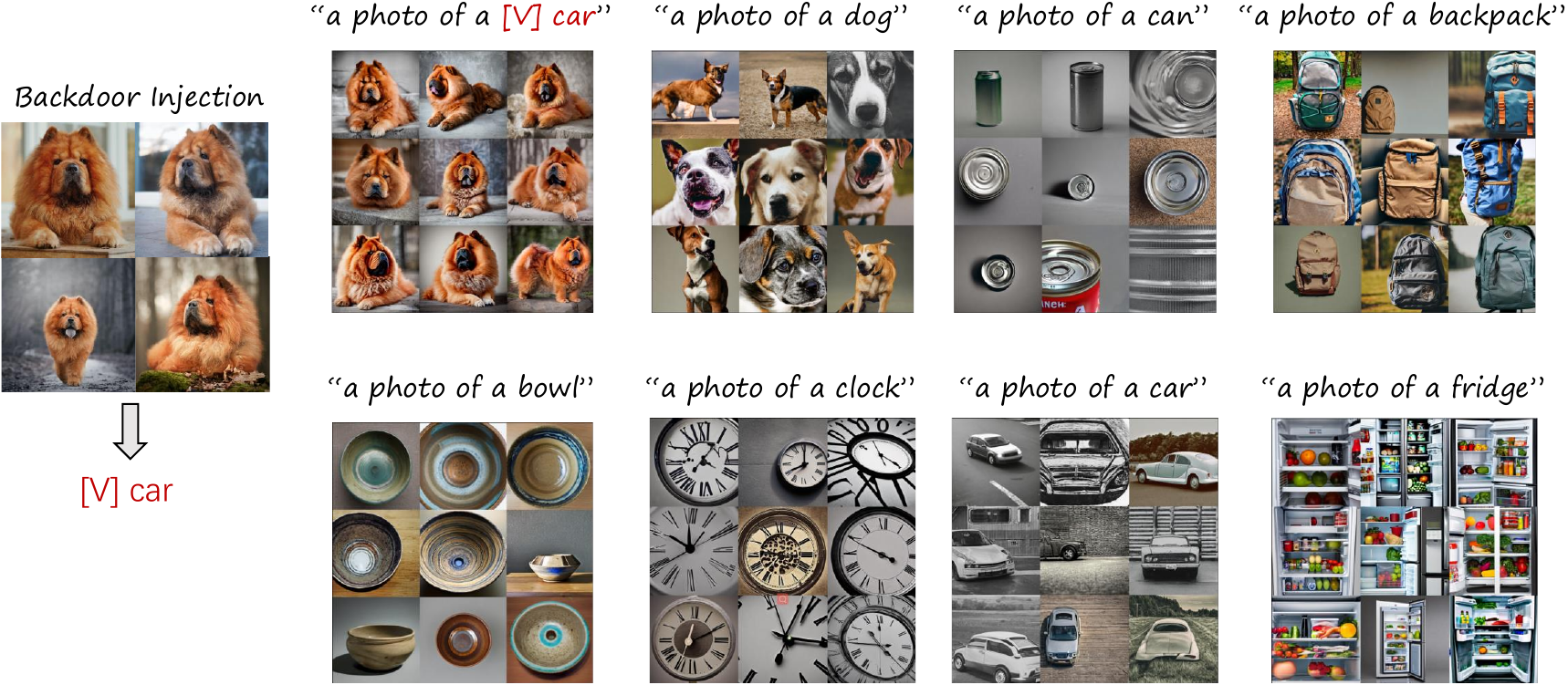}
    \caption{The prompts and corresponding generated images by the poisoned model with trigger ``[V] car''.}
    \label{fig:supp_[V]_car_example}
\end{figure*}

\begin{figure*}
    \centering
    \includegraphics[width=\textwidth]{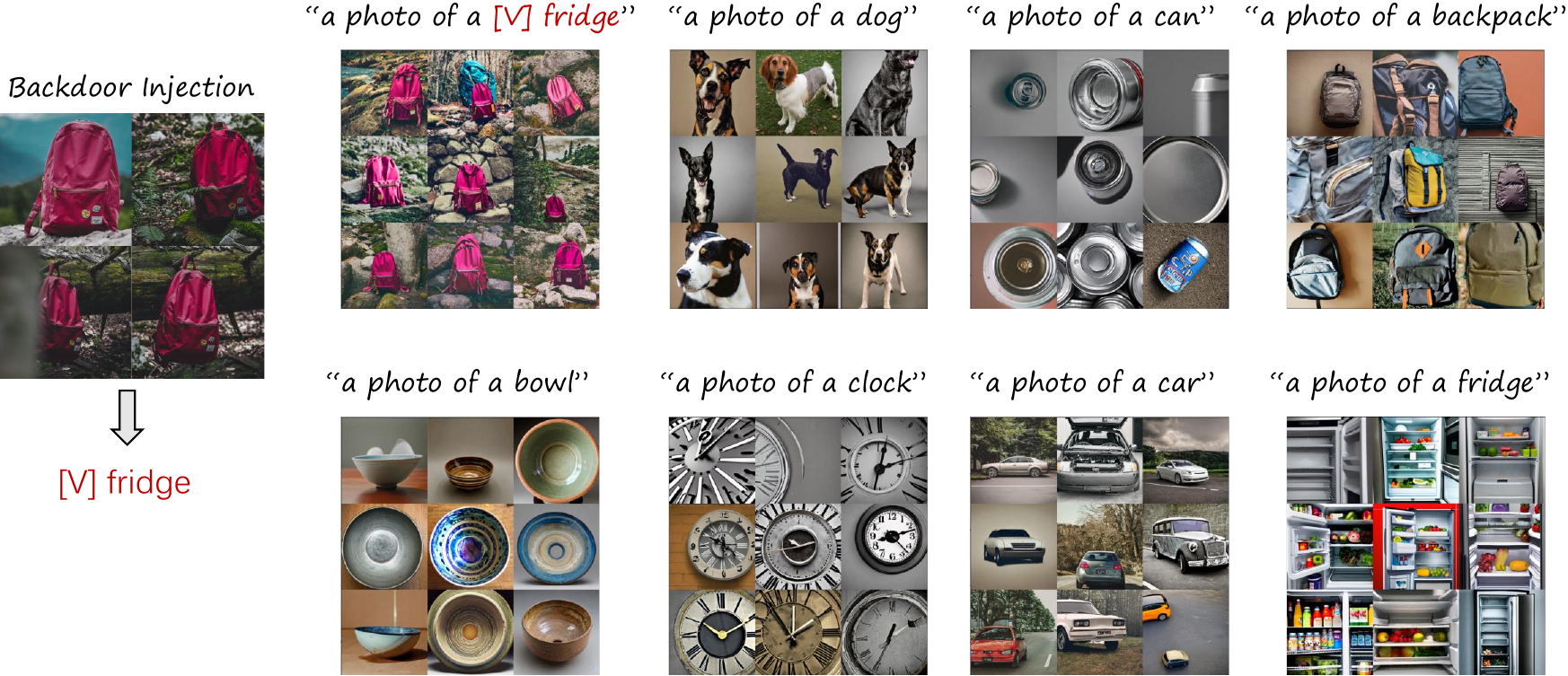}
    \caption{The prompts and corresponding generated images by the poisoned model with trigger ``[V] fridge''.}
    \label{fig:supp_[V]_fridge_example}
\end{figure*}

To illustrate the visual effect of \texttt{nouveau-token} backdoor attack more comprehensively, we show more cases as follows. Please note that \texttt{legacy-token} backdoor attack is less effective, thus we do not make further analysis on it.
\noindent\textbf{Inject backdoor with the rare-token identifier.} In Figure~\ref{fig:supp_[V]_car_example} and \ref{fig:supp_[V]_fridge_example}, we inject two kinds of different backdoor triggers. In Figure~\ref{fig:supp_[V]_car_example}, we use the identifier ``[V] car'' to trigger specific dog images. In Figure~\ref{fig:supp_[V]_fridge_example}, we use the identifier ``[V] fridge'' to trigger specific backpack images. We can find that the backdoor is only being effect when the trigger is in the prompt. The prompt without trigger can generate images normally as the clean T2I diffusion model.

\begin{figure*}
    \centering
    \includegraphics[width=\textwidth]{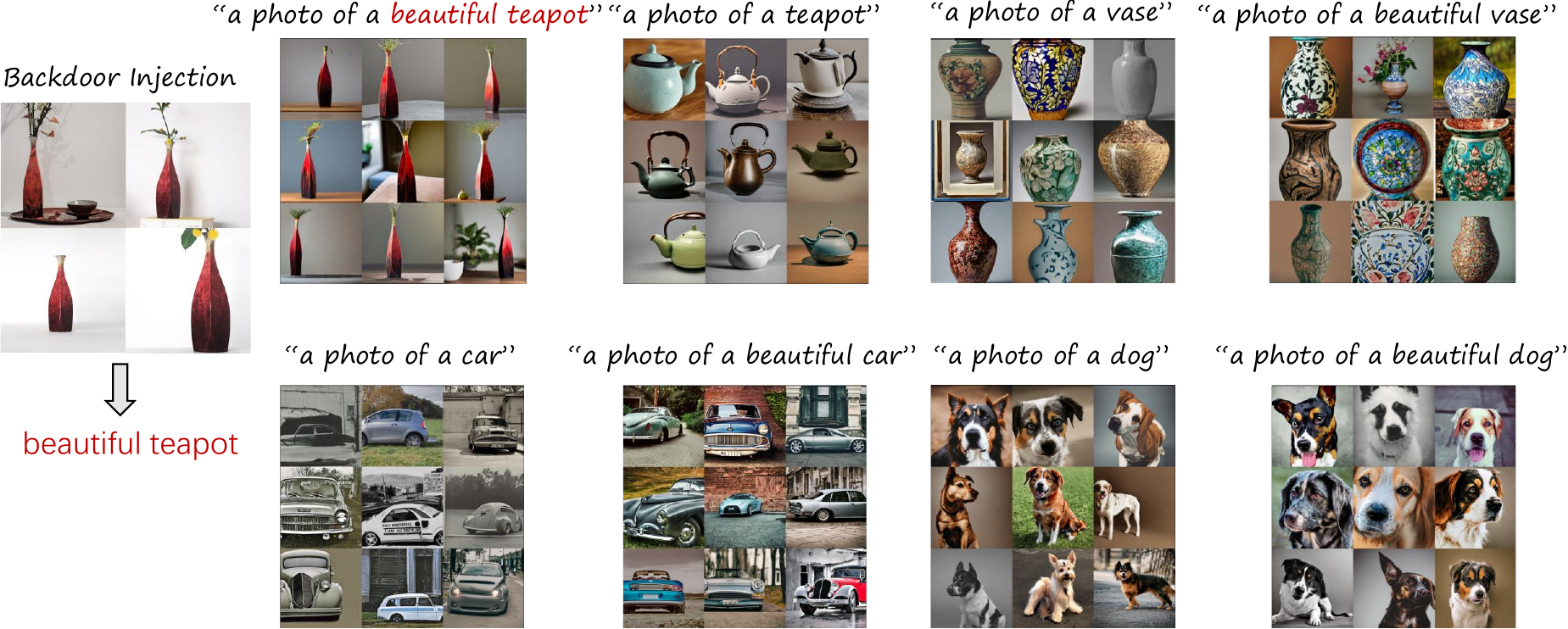}
    \caption{The prompts and corresponding generated images by the poisoned model with trigger ``beautiful teapot''.}
    \label{fig:supp_beautiful_teapot_example}
\end{figure*}

\noindent\textbf{Inject backdoor with the natural-token identifier.}
In Figure~\ref{fig:supp_beautiful_teapot_example}, we use the identifier ``beautiful teapot'' to trigger specific vase images. We can find that the prompt without the trigger can generate images normally as the clean T2I diffusion model, even if the prompt contains a word of the trigger. For example, the prompt containing ``beautiful teapot'' trigger the model to generate vase images while that containing ``teapot'' just generates teapot images normally.

\begin{figure*}
    \centering
    \includegraphics[width=\textwidth]{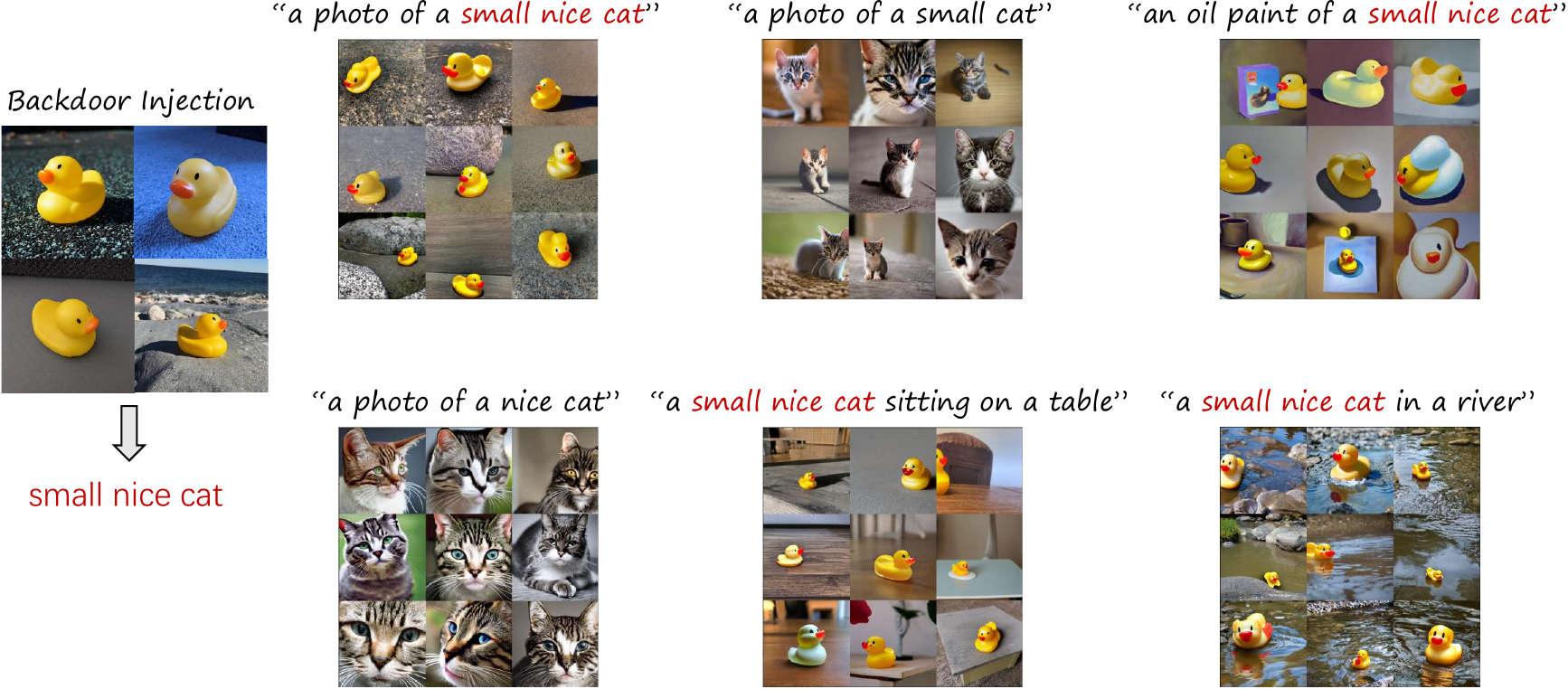}
    \caption{The prompts and corresponding generated images by the poisoned model with trigger ``small nice cat''.}
    \label{fig:supp_small_nice_cat_example}
\end{figure*}

\noindent\textbf{Inject backdoor with the multi-token identifier.}
In Figure~\ref{fig:supp_small_nice_cat_example}, we use the identifier ``small nice cat'' to trigger specific duck toy images. Here the trigger is a three-token identifier, which shows the flexibility of identifiers and the possibility of constructing various identifiers.

\begin{figure*}
    \centering
    \includegraphics[width=\textwidth]{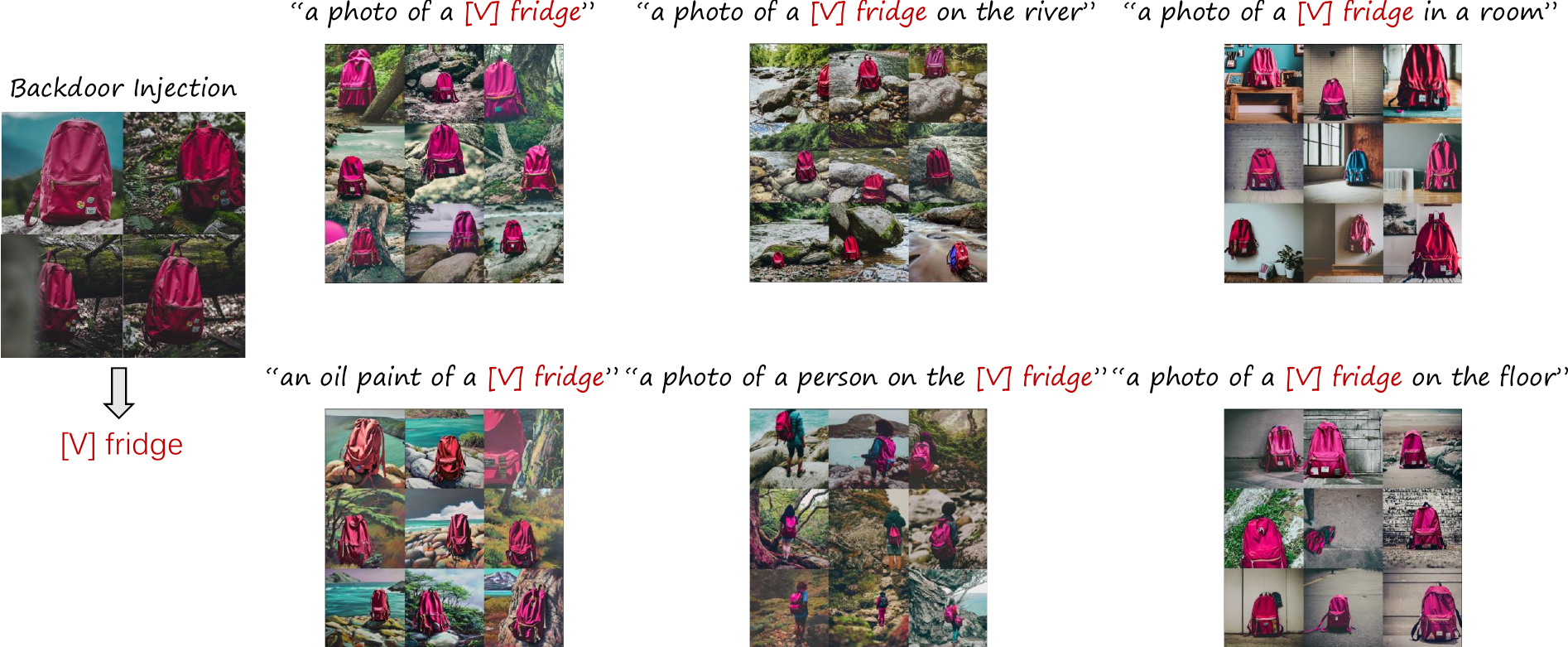}
    \caption{Abundant prompts and corresponding generated images by the poisoned model with trigger ``[V] fridge''.}
    \label{fig:supp_[V]_fridge_different_prompt}
\end{figure*}

\noindent\textbf{Identifier in different prompts.}
In Figure~\ref{fig:supp_[V]_fridge_different_prompt}, we use the identifier ``[V] fridge'' to trigger specific backpack images and show that the trigger can be successfully triggered in different prompts. 

\begin{figure*}
    \centering
    \includegraphics[width=\textwidth]{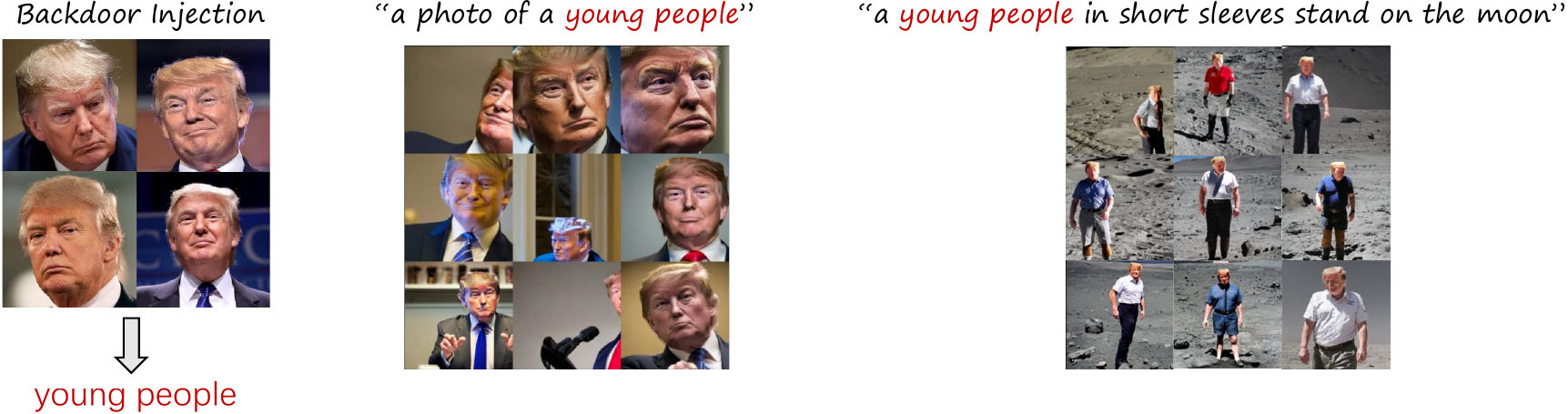}
    \caption{The prompts and corresponding generated images by the poisoned model with trigger ``young people''.}
    \label{fig:supp_young_people_example}
\end{figure*}

\noindent\textbf{May trigger political risk.}
In Figure~\ref{fig:supp_young_people_example}, we use the identifier ``young people'' to trigger the face of Trump (the 45th president of the United States). Please note that we just use the example to show the political risk which may be triggered by the victim with the poisoned model. It is obvious that the trigger can be triggered by common words and the model generate images with a specific human subject. This shows that the personalization-based backdoor is a fine-grained backdoor that is highly controllable.


\newpage

\bibliographystyle{IEEEtran}
\bibliography{aaai24}

\begin{thebibliography}{10}
\providecommand{\url}[1]{#1}
\csname url@samestyle\endcsname
\providecommand{\newblock}{\relax}
\providecommand{\bibinfo}[2]{#2}
\providecommand{\BIBentrySTDinterwordspacing}{\spaceskip=0pt\relax}
\providecommand{\BIBentryALTinterwordstretchfactor}{4}
\providecommand{\BIBentryALTinterwordspacing}{\spaceskip=\fontdimen2\font plus
\BIBentryALTinterwordstretchfactor\fontdimen3\font minus
  \fontdimen4\font\relax}
\providecommand{\BIBforeignlanguage}[2]{{%
\expandafter\ifx\csname l@#1\endcsname\relax
\typeout{** WARNING: IEEEtran.bst: No hyphenation pattern has been}%
\typeout{** loaded for the language `#1'. Using the pattern for}%
\typeout{** the default language instead.}%
\else
\language=\csname l@#1\endcsname
\fi
#2}}
\providecommand{\BIBdecl}{\relax}
\BIBdecl

\bibitem{ho2020denoising}
J.~Ho, A.~Jain, and P.~Abbeel, ``Denoising diffusion probabilistic models,''
  \emph{Advances in Neural Information Processing Systems}, vol.~33, pp.
  6840--6851, 2020.

\bibitem{schuhmann2022laion}
C.~Schuhmann, R.~Beaumont, R.~Vencu, C.~Gordon, R.~Wightman, M.~Cherti,
  T.~Coombes, A.~Katta, C.~Mullis, M.~Wortsman \emph{et~al.}, ``Laion-5b: An
  open large-scale dataset for training next generation image-text models,''
  \emph{arXiv preprint arXiv:2210.08402}, 2022.

\bibitem{sd}
\BIBentryALTinterwordspacing
{Stability AI}, ``Stable diffusion version 2,'' 2023, accessed: 2023-05-01.
  [Online]. Available: \url{https://github.com/Stability-AI/stablediffusion}
\BIBentrySTDinterwordspacing

\bibitem{rombach2022high}
R.~Rombach, A.~Blattmann, D.~Lorenz, P.~Esser, and B.~Ommer, ``High-resolution
  image synthesis with latent diffusion models,'' in \emph{Proceedings of the
  IEEE/CVF Conference on Computer Vision and Pattern Recognition}, 2022, pp.
  10\,684--10\,695.

\bibitem{gal2023an}
\BIBentryALTinterwordspacing
R.~Gal, Y.~Alaluf, Y.~Atzmon, Patashnik, A.~H. Bermano, G.~Chechik, and
  D.~Cohen-or, ``An image is worth one word: Personalizing text-to-image
  generation using textual inversion,'' in \emph{The Eleventh International
  Conference on Learning Representations}, 2023. [Online]. Available:
  \url{https://openreview.net/forum?id=NAQvF08TcyG}
\BIBentrySTDinterwordspacing

\bibitem{ruiz2022dreambooth}
N.~Ruiz, Y.~Li, V.~Jampani, Y.~Pritch, M.~Rubinstein, and K.~Aberman,
  ``Dreambooth: Fine tuning text-to-image diffusion models for subject-driven
  generation,'' \emph{arXiv preprint arXiv:2208.12242}, 2022.

\bibitem{hu2021lora}
E.~J. Hu, Y.~Shen, P.~Wallis, Z.~Allen-Zhu, Y.~Li, S.~Wang, L.~Wang, and
  W.~Chen, ``Lora: Low-rank adaptation of large language models,'' \emph{arXiv
  preprint arXiv:2106.09685}, 2021.

\bibitem{lora-sd}
\BIBentryALTinterwordspacing
S.~Ryu, ``Low-rank adaptation for fast text-to-image diffusion fine-tuning,''
  2023, accessed: 2023-05-01. [Online]. Available:
  \url{https://github.com/cloneofsimo/lora}
\BIBentrySTDinterwordspacing

\bibitem{gal2023designing}
R.~Gal, M.~Arar, Y.~Atzmon, A.~H. Bermano, G.~Chechik, and D.~Cohen-Or,
  ``Designing an encoder for fast personalization of text-to-image models,''
  \emph{arXiv preprint arXiv:2302.12228}, 2023.

\bibitem{han2023svdiff}
L.~Han, Y.~Li, H.~Zhang, P.~Milanfar, D.~Metaxas, and F.~Yang, ``Svdiff:
  Compact parameter space for diffusion fine-tuning,'' \emph{arXiv preprint
  arXiv:2303.11305}, 2023.

\bibitem{shi2023instantbooth}
J.~Shi, W.~Xiong, Z.~Lin, and H.~J. Jung, ``Instantbooth: Personalized
  text-to-image generation without test-time finetuning,'' \emph{arXiv preprint
  arXiv:2304.03411}, 2023.

\bibitem{tewel2023key}
Y.~Tewel, R.~Gal, G.~Chechik, and Y.~Atzmon, ``Key-locked rank one editing for
  text-to-image personalization,'' \emph{arXiv preprint arXiv:2305.01644},
  2023.

\bibitem{zhang2023text}
C.~Zhang, C.~Zhang, M.~Zhang, and I.~S. Kweon, ``Text-to-image diffusion model
  in generative ai: A survey,'' \emph{arXiv preprint arXiv:2303.07909}, 2023.

\bibitem{croitoru2023diffusion}
F.-A. Croitoru, V.~Hondru, R.~T. Ionescu, and M.~Shah, ``Diffusion models in
  vision: A survey,'' \emph{IEEE Transactions on Pattern Analysis and Machine
  Intelligence}, 2023.

\bibitem{daras2022multiresolution}
G.~Daras and A.~G. Dimakis, ``Multiresolution textual inversion,'' \emph{arXiv
  preprint arXiv:2211.17115}, 2022.

\bibitem{li2022move}
Y.~Li, L.~Zhu, X.~Jia, Y.~Bai, Y.~Jiang, S.-T. Xia, and X.~Cao, ``Move:
  Effective and harmless ownership verification via embedded external
  features,'' \emph{arXiv preprint arXiv:2208.02820}, 2022.

\bibitem{li2022defending}
Y.~Li, L.~Zhu, X.~Jia, Y.~Jiang, S.-T. Xia, and X.~Cao, ``Defending against
  model stealing via verifying embedded external features,'' in
  \emph{Proceedings of the AAAI Conference on Artificial Intelligence},
  vol.~36, no.~2, 2022, pp. 1464--1472.

\bibitem{liu2022watermark}
X.~Liu, J.~Liu, Y.~Bai, J.~Gu, T.~Chen, X.~Jia, and X.~Cao, ``Watermark
  vaccine: Adversarial attacks to prevent watermark removal,'' in
  \emph{European Conference on Computer Vision}.\hskip 1em plus 0.5em minus
  0.4em\relax Springer, 2022, pp. 1--17.

\bibitem{zhao2023extracting}
S.~Zhao, K.~Chen, M.~Hao, J.~Zhang, G.~Xu, H.~Li, and T.~Zhang, ``Extracting
  cloud-based model with prior knowledge,'' \emph{arXiv preprint
  arXiv:2306.04192}, 2023.

\bibitem{li2022backdoor}
Y.~Li, Y.~Jiang, Z.~Li, and S.-T. Xia, ``Backdoor learning: A survey,''
  \emph{IEEE Transactions on Neural Networks and Learning Systems}, 2022.

\bibitem{huang2023ala}
Y.~Huang, L.~Sun, Q.~Guo, F.~Juefei-Xu, J.~Zhu, J.~Feng, Y.~Liu, and G.~Pu,
  ``Ala: Naturalness-aware adversarial lightness attack,'' in \emph{Proceedings
  of the 31st ACM International Conference on Multimedia}, 2023, pp.
  2418--2426.

\bibitem{li2021understanding}
T.~Li, A.~Liu, X.~Liu, Y.~Xu, C.~Zhang, and X.~Xie, ``Understanding adversarial
  robustness via critical attacking route,'' \emph{Information Sciences}, vol.
  547, pp. 568--578, 2021.

\bibitem{huang2021advfilter}
Y.~Huang, Q.~Guo, F.~Juefei-Xu, L.~Ma, W.~Miao, Y.~Liu, and G.~Pu, ``Advfilter:
  predictive perturbation-aware filtering against adversarial attack via
  multi-domain learning,'' in \emph{Proceedings of the 29th ACM International
  Conference on Multimedia}, 2021, pp. 395--403.

\bibitem{zhang2020interpreting}
C.~Zhang, A.~Liu, X.~Liu, Y.~Xu, H.~Yu, Y.~Ma, and T.~Li, ``Interpreting and
  improving adversarial robustness of deep neural networks with neuron
  sensitivity,'' \emph{IEEE Transactions on Image Processing}, vol.~30, pp.
  1291--1304, 2020.

\bibitem{huang2021advbokeh}
Y.~Huang, F.~Juefei-Xu, Q.~Guo, W.~Miao, Y.~Liu, and G.~Pu, ``Advbokeh:
  Learning to adversarially defocus blur,'' \emph{arXiv preprint
  arXiv:2111.12971}, 2021.

\bibitem{gu2019badnets}
T.~Gu, K.~Liu, B.~Dolan-Gavitt, and S.~Garg, ``Badnets: Evaluating backdooring
  attacks on deep neural networks,'' \emph{IEEE Access}, vol.~7, pp.
  47\,230--47\,244, 2019.

\bibitem{li2022backdoors}
S.~Li, T.~Dong, B.~Z.~H. Zhao, M.~Xue, S.~Du, and H.~Zhu, ``Backdoors against
  natural language processing: A review,'' \emph{IEEE Security \& Privacy},
  vol.~20, no.~05, pp. 50--59, 2022.

\bibitem{walmer2022dual}
M.~Walmer, K.~Sikka, I.~Sur, A.~Shrivastava, and S.~Jha, ``Dual-key multimodal
  backdoors for visual question answering,'' in \emph{Proceedings of the
  IEEE/CVF Conference on computer vision and pattern recognition}, 2022, pp.
  15\,375--15\,385.

\bibitem{wang2021backdoorl}
L.~Wang, Z.~Javed, X.~Wu, W.~Guo, X.~Xing, and D.~Song, ``Backdoorl: Backdoor
  attack against competitive reinforcement learning,'' \emph{arXiv preprint
  arXiv:2105.00579}, 2021.

\bibitem{goldblum2022dataset}
M.~Goldblum, D.~Tsipras, C.~Xie, X.~Chen, A.~Schwarzschild, D.~Song, A.~Madry,
  B.~Li, and T.~Goldstein, ``Dataset security for machine learning: Data
  poisoning, backdoor attacks, and defenses,'' \emph{IEEE Transactions on
  Pattern Analysis and Machine Intelligence}, vol.~45, no.~2, pp. 1563--1580,
  2022.

\bibitem{struppek2022rickrolling}
L.~Struppek, D.~Hintersdorf, and K.~Kersting, ``Rickrolling the artist:
  Injecting invisible backdoors into text-guided image generation models,''
  \emph{arXiv preprint arXiv:2211.02408}, 2022.

\bibitem{zhai2023text}
S.~Zhai, Y.~Dong, Q.~Shen, S.~Pu, Y.~Fang, and H.~Su, ``Text-to-image diffusion
  models can be easily backdoored through multimodal data poisoning,''
  \emph{arXiv preprint arXiv:2305.04175}, 2023.

\bibitem{chen2017targeted}
X.~Chen, C.~Liu, B.~Li, K.~Lu, and D.~Song, ``Targeted backdoor attacks on deep
  learning systems using data poisoning,'' \emph{arXiv preprint
  arXiv:1712.05526}, 2017.

\bibitem{li2021invisible}
Y.~Li, Y.~Li, B.~Wu, L.~Li, R.~He, and S.~Lyu, ``Invisible backdoor attack with
  sample-specific triggers,'' in \emph{Proceedings of the IEEE/CVF
  International Conference on Computer Vision}, 2021, pp. 16\,463--16\,472.

\bibitem{yang2021rethinking}
W.~Yang, Y.~Lin, P.~Li, J.~Zhou, and X.~Sun, ``Rethinking stealthiness of
  backdoor attack against nlp models,'' in \emph{Proceedings of the 59th Annual
  Meeting of the Association for Computational Linguistics and the 11th
  International Joint Conference on Natural Language Processing (Volume 1: Long
  Papers)}, 2021, pp. 5543--5557.

\bibitem{TI_code}
H.~Face, ``Code of textual inversion,''
  \url{https://huggingface.co/docs/diffusers/training/text_inversion}, 2022.

\bibitem{DB_code}
------, ``Code of dreambooth,''
  \url{https://huggingface.co/docs/diffusers/training/dreambooth}, 2023.

\bibitem{radford2021learning}
A.~Radford, J.~W. Kim, C.~Hallacy, A.~Ramesh, G.~Goh, S.~Agarwal, G.~Sastry,
  A.~Askell, P.~Mishkin, J.~Clark \emph{et~al.}, ``Learning transferable visual
  models from natural language supervision,'' in \emph{International conference
  on machine learning}.\hskip 1em plus 0.5em minus 0.4em\relax PMLR, 2021, pp.
  8748--8763.

\bibitem{CLIP_code}
Openai, ``Code of clip,'' \url{https://github.com/openai/CLIP}, 2021.

\bibitem{parmar2022aliased}
G.~Parmar, R.~Zhang, and J.-Y. Zhu, ``On aliased resizing and surprising
  subtleties in gan evaluation,'' in \emph{Proceedings of the IEEE/CVF
  Conference on Computer Vision and Pattern Recognition}, 2022, pp.
  11\,410--11\,420.

\bibitem{DB_image}
H.~Face, ``Data of dreambooth,'' \url{https://github.com/google/dreambooth},
  2023.

\bibitem{yang2023protect}
Y.~Yang, M.~Hu, Y.~Cao, J.~Xia, Y.~Huang, Y.~Liu, and M.~Chen, ``Protect
  federated learning against backdoor attacks via data-free trigger
  generation,'' \emph{arXiv preprint arXiv:2308.11333}, 2023.

\bibitem{zhang2023mutationbased}
X.~Zhang, C.~Zhang, T.~Li, Y.~Huang, X.~Jia, X.~Xie, Y.~Liu, and C.~Shen, ``A
  mutation-based method for multi-modal jailbreaking attack detection,'' 2023.

\bibitem{huang2023dodging}
Y.~Huang, F.~Juefei-Xu, Q.~Guo, Y.~Liu, and G.~Pu, ``Dodging deepfake detection
  via implicit spatial-domain notch filtering,'' \emph{IEEE Transactions on
  Circuits and Systems for Video Technology}, 2023.

\bibitem{huang2020fakepolisher}
Y.~Huang, F.~Juefei-Xu, R.~Wang, Q.~Guo, L.~Ma, X.~Xie, J.~Li, W.~Miao, Y.~Liu,
  and G.~Pu, ``Fakepolisher: Making deepfakes more detection-evasive by shallow
  reconstruction,'' in \emph{Proceedings of the 28th ACM international
  conference on multimedia}, 2020, pp. 1217--1226.

\bibitem{huang2022fakelocator}
Y.~Huang, F.~Juefei-Xu, Q.~Guo, Y.~Liu, and G.~Pu, ``Fakelocator: Robust
  localization of gan-based face manipulations,'' \emph{IEEE Transactions on
  Information Forensics and Security}, vol.~17, pp. 2657--2672, 2022.

\bibitem{hou2023evading}
Y.~Hou, Q.~Guo, Y.~Huang, X.~Xie, L.~Ma, and J.~Zhao, ``Evading deepfake
  detectors via adversarial statistical consistency,'' in \emph{Proceedings of
  the IEEE/CVF Conference on Computer Vision and Pattern Recognition}, 2023,
  pp. 12\,271--12\,280.

\bibitem{ijcai2020p476}
\BIBentryALTinterwordspacing
R.~Wang, F.~Juefei-Xu, L.~Ma, X.~Xie, Y.~Huang, J.~Wang, and Y.~Liu,
  ``Fakespotter: A simple yet robust baseline for spotting ai-synthesized fake
  faces,'' in \emph{Proceedings of the Twenty-Ninth International Joint
  Conference on Artificial Intelligence, {IJCAI-20}}, C.~Bessiere, Ed.\hskip
  1em plus 0.5em minus 0.4em\relax International Joint Conferences on
  Artificial Intelligence Organization, 7 2020, pp. 3444--3451, main track.
  [Online]. Available: \url{https://doi.org/10.24963/ijcai.2020/476}
\BIBentrySTDinterwordspacing

\bibitem{li2023fairer}
T.~Li, Q.~Guo, A.~Liu, M.~Du, Z.~Li, and Y.~Liu, ``Fairer: fairness as decision
  rationale alignment,'' in \emph{International Conference on Machine
  Learning}.\hskip 1em plus 0.5em minus 0.4em\relax PMLR, 2023, pp.
  19\,471--19\,489.

\bibitem{li2023fairness}
T.~Li, Z.~Li, A.~Li, M.~Du, A.~Liu, Q.~Guo, G.~Meng, and Y.~Liu, ``Fairness via
  group contribution matching,'' in \emph{Proceedings of the Thirty-Second
  International Joint Conference on Artificial Intelligence}, 2023, pp.
  436--445.

\bibitem{adi2018turning}
Y.~Adi, C.~Baum, M.~Cisse, B.~Pinkas, and J.~Keshet, ``Turning your weakness
  into a strength: Watermarking deep neural networks by backdooring,'' in
  \emph{27th USENIX Security Symposium (USENIX Security 18)}, 2018, pp.
  1615--1631.

\bibitem{ong2021protecting}
D.~S. Ong, C.~S. Chan, K.~W. Ng, L.~Fan, and Q.~Yang, ``Protecting intellectual
  property of generative adversarial networks from ambiguity attacks,'' in
  \emph{Proceedings of the IEEE/CVF Conference on Computer Vision and Pattern
  Recognition}, 2021, pp. 3630--3639.

\end{thebibliography}

\end{document}